\documentclass{article}

\usepackage{arxiv}

\usepackage[utf8]{inputenc} % allow utf-8 input
\usepackage[T1]{fontenc}    % use 8-bit T1 fonts
\usepackage{hyperref}       % hyperlinks
\usepackage{url}            % simple URL typesetting
\usepackage{booktabs}       % professional-quality tables
\usepackage{amsfonts}       % blackboard math symbols
\usepackage{amssymb}
\usepackage{amsthm}
\usepackage{nicefrac}       % compact symbols for 1/2, etc.
\usepackage{microtype}      % microtypography
\usepackage{lipsum}
\usepackage{graphicx}

\usepackage{bbm}
\usepackage{paralist}
\usepackage[ruled,linesnumbered]{algorithm2e}

\setcitestyle{square,numbers,comma, sort&compress}
\usepackage{bbm}
\usepackage{paralist}
\usepackage[ruled,linesnumbered]{algorithm2e}
\usepackage{diagbox}
\usepackage{array}
\newcolumntype{P}[1]{>{\centering\arraybackslash}p{#1}}
\usepackage{times}
\usepackage{epsfig}
\usepackage{graphicx}
\usepackage{amsmath}
\usepackage{amssymb}
\usepackage{color}
\usepackage{amsmath}
\usepackage{comment}

\usepackage[flushmargin]{footmisc} 

\specialcomment{notincluded}{}{}
\excludecomment{notincluded}

\DeclareMathOperator*{\argmin}{argmin}

\newcommand{\etal}{\textit{et al}. }

\renewcommand{\footnoterule}{%
  \hrule width 0pt height 0pt
  \kern 25pt
}

\let\svthefootnote\thefootnote
\newcommand\blankfootnote[1]{%

  \let\thefootnote\relax\footnotetext{#1}%
  \let\thefootnote\svthefootnote%
}

\title{Unsupervised Meta-Learning for Few-Shot Image Classification}

\author{Siavash Khodadadeh, Ladislau B{\"o}l{\"o}ni\\
Dept. of Computer Science\\
University of Central Florida\\
{\tt\small siavash.khodadadeh@knights.ucf.edu, lboloni@cs.ucf.edu}
\And
Mubarak Shah\\
Center for Research in Computer Vision\\
University of Central Florida\\
{\tt\small shah@crcv.ucf.edu}
}

\begin{document}
\maketitle

\begin{abstract}

 Few-shot or one-shot learning of classifiers requires a significant inductive bias towards the type of task to be learned. One way to acquire this is by meta-learning on tasks similar to the target task. 
 In this paper, we propose UMTRA, an algorithm that performs unsupervised, model-agnostic meta-learning for classification tasks. 
 
 The meta-learning step of UMTRA is performed on a flat collection of unlabeled images. While we assume that these images can be grouped into a diverse set of classes and are relevant to the target task, no explicit information about the classes or any labels are needed. UMTRA uses random sampling and augmentation to create synthetic training tasks for meta-learning phase. Labels are only needed at the final target task learning step, and they can be as little as one sample per class.

On the Omniglot and Mini-Imagenet few-shot learning benchmarks, UMTRA outperforms every tested approach based on unsupervised learning of representations, while alternating for the best performance with the recent CACTUs algorithm. Compared to supervised model-agnostic meta-learning approaches, UMTRA trades off some classification accuracy for a reduction in the required labels of several orders of magnitude. 

\end{abstract}

\blankfootnote{33rd Conference on Neural Information Processing Systems (NeurIPS 2019), Vancouver, Canada.}

\section{Introduction}
\label{sec:Introduction}

Meta-learning or ``learning-to-learn'' approaches have been proposed in the neural networks literature since the 1980s~\cite{schmidhuber1987evolutionary, bengio1990learning}. The general idea is to prepare the network through several 
% auxiliary 
learning tasks $\mathcal{T}_1 \ldots \mathcal{T}_n$, in a {\em meta-learning phase} such that when presented with the target task $\mathcal{T}_{n + 1}$, the network will be ready to learn it as efficiently as possible. 

Recently proposed model-agnostic meta-learning approaches \cite{finn2017model,Nichol-2018-Reptile} can be applied to any differentiable network. When used for classification, the target learning phase consists of several gradient descent steps on a backpropagated supervised classification loss. Unfortunately, these approaches require the 
% auxiliary 
learning tasks $\mathcal{T}_i$ to have the same supervised learning format as the target task. Acquiring labeled data for a large number of tasks is not only a problem of cost and convenience but also puts conceptual limits on the type of problems that can be solved through meta-learning. If we need to have labeled training data for tasks $\mathcal{T}_1 \ldots \mathcal{T}_n$ in order to learn task $\mathcal{T}_{n+1}$, this limits us to task types that are variations of tasks known and solved (at least by humans). 

\begin{figure}
\centering
Supervised MAML \\ % \hspace{5cm} UMTRA
\includegraphics[width=0.75\columnwidth]{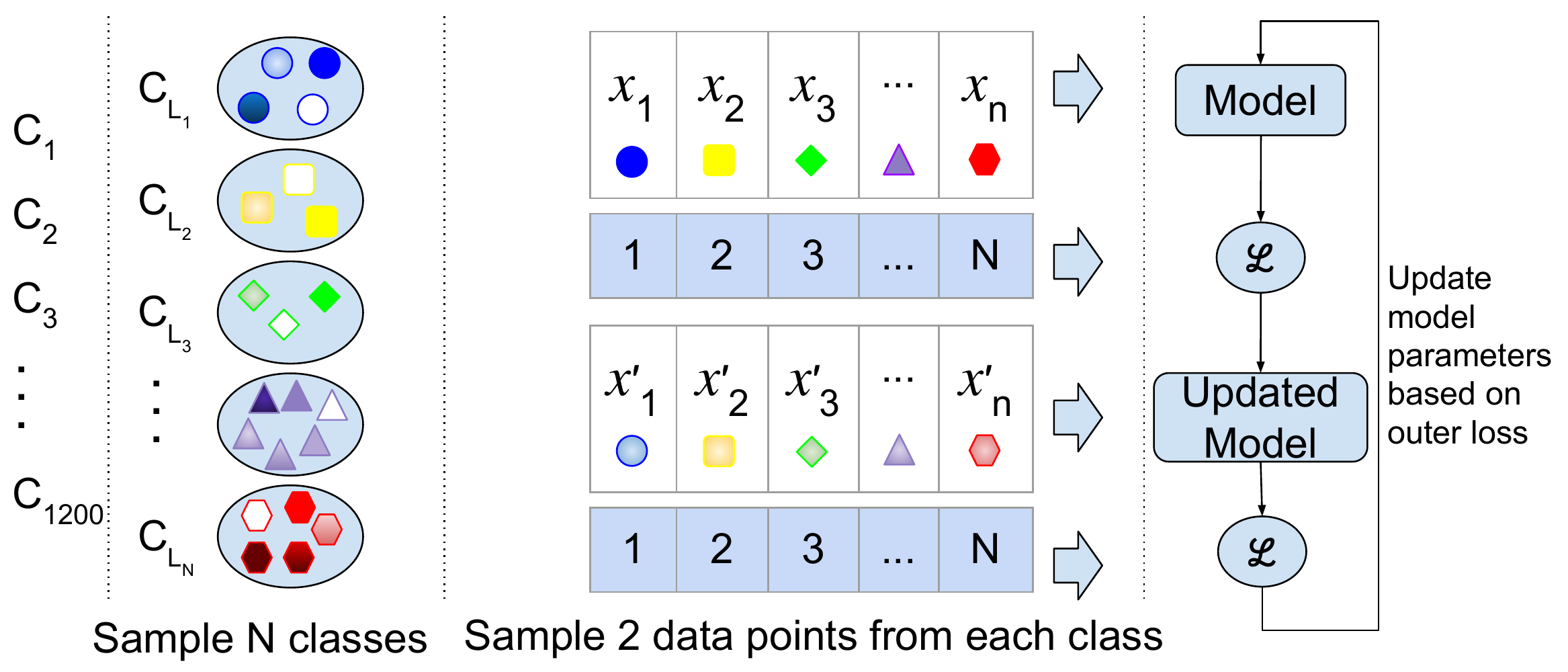} \\
\vspace{0.5cm}
UMTRA \\
\includegraphics[width=0.75\columnwidth]{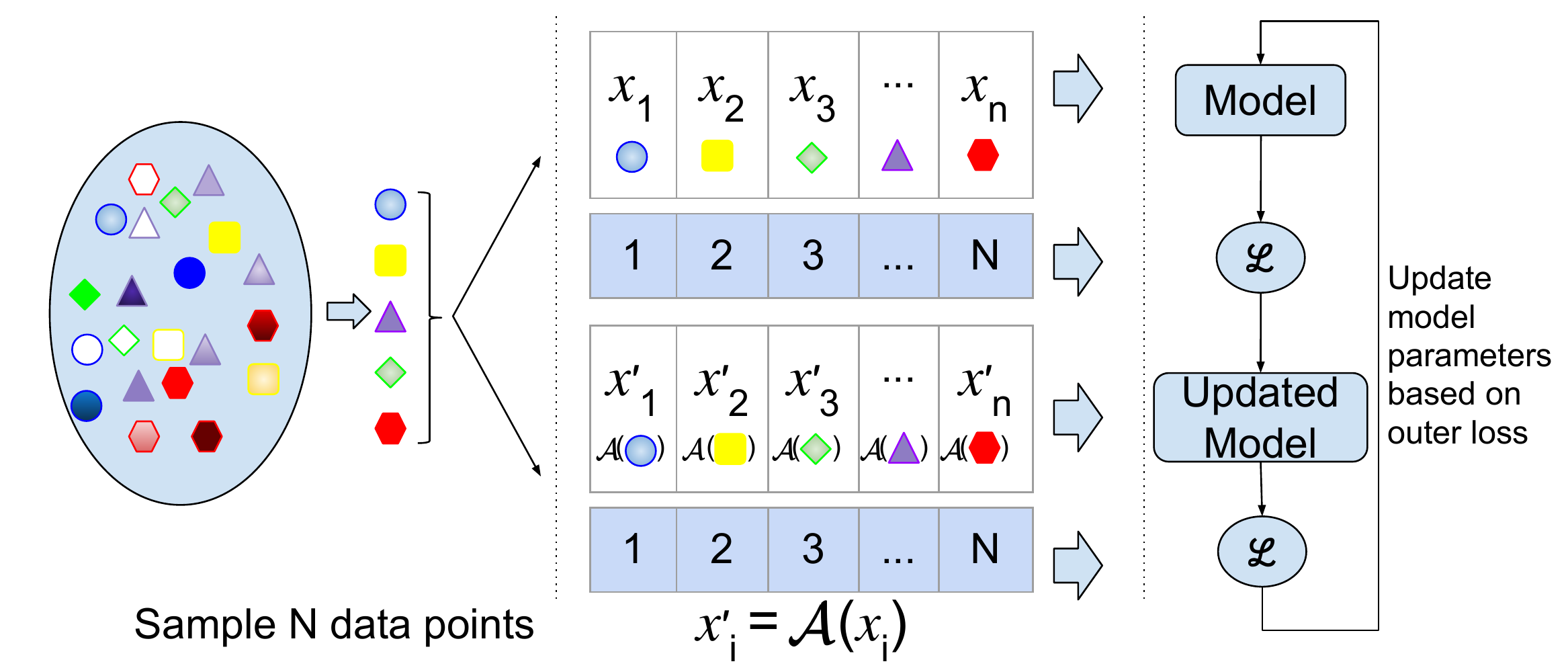}
\caption{The process of creation of the training and validation data of the meta-training task $\mathcal{T}$. (top) Supervised MAML: We start from a dataset where the samples are labeled with their class. The training data is created by sampling $N$ distinct classes $C_{L_i}$, and choosing a random sample $x_i$ from each. The validation data is created by choosing a different sample $x_i'$ from the same class. (bottom) UMTRA: We start from a dataset of unlabeled data. The training data is created by randomly choosing N samples $x_i$ from the dataset. 
The validation data is created by applying the augmentation function $\mathcal{A}$ to each sample from the training data. For both MAML and UMTRA, artificial temporary labels $1, 2 \ldots N$ are used. }
\label{fig:story}
\end{figure}

In this paper, we propose an algorithm called Unsupervised Meta-learning with Tasks constructed by Random sampling and Augmentation (UMTRA) that performs meta-learning of one-shot or few-shot classifiers in an unsupervised manner on an unlabeled dataset. Instead of starting from a collection of labeled tasks, $\{\ldots \mathcal{T}_i \ldots\}$, UMTRA starts with a collection of unlabeled data $\mathcal{U}=\{\ldots x_i \ldots\}$. We have only a set of relatively easy-to-satisfy requirements towards $\mathcal{U}$: Its objects have to be drawn from the same distribution as the objects classified in the target task and it must have a set of classes significantly larger than the number of classes of the final classifier. Starting from this unlabeled dataset, UMTRA uses statistical diversity properties and domain specific augmentations to generate the training and validation data for a collection of synthetic tasks, $\{\ldots \mathcal{T}_i' \ldots\}$. These tasks are then used in the meta-learning process based on a modified classification variant of the MAML algorithm~\cite{finn2017model}. Figure~\ref{fig:story} summarizes the differences between the original supervised MAML model and the process of generating synthetic tasks from unsupervised data in UMTRA. 

The contributions of this paper can be summarized as follows:

\begin{compactitem}

    \item We describe a novel algorithm that allows unsupervised, model-agnostic meta-learning for few-shot classification by generating synthetic meta-learning data with artificial labels. 
    % In recent months, two algorithms proposing similar ideas of generating synthetic training data from unsupervised data were proposed: CACTUs~\cite{hsu2018unsupervised} and AAL~\cite{antoniou2019assume}. 
    
    \item From a theoretical point of view, we demonstrate a relationship between generalization error and the loss backpropagated from the validation set in MAML. Our intuition is that we can generate unsupervised validation tasks which can perform effectively if we are able to span the space of the classes by generating useful samples with augmentation.

    \item
    On all the Omniglot and Mini-Imagenet few-shot learning benchmarks, UMTRA outperforms every tested approach based on unsupervised learning of representations. It also achieves a significant percentage of the accuracy of the supervised MAML approach, while requiring vastly fewer labels. 
    % For instance, on five-way one-shot Omniglot classification, our approach retains 89\% of the classification accuracy of MAML while reducing the number of required labels from {\bf 24005 to 5}.
    For instance, for 5-way 5-shot classification on the Omniglot dataset UMTRA obtains a 95.43\% accuracy with only {\bf 25} labels, while supervised MAML obtains 98.83\% with 24025. 
    Compared with recent unsupervised meta-learning approaches building on top of stock MAML, UMTRA alternates for the best performance with the CACTUs  algorithm. 

    % \item We demonstrate the ability to perform one-shot video classification using meta-learning (both supervised and unsupervised). We show that the algorithm outperforms pre-training based approaches, and works even when the meta-learning database (Kinetics) is different from the test database (UCF-101). To the  best of our knowledge, this is the first work to show {\em any} type of meta-learning for video classification.
    
    % \item \MS{how about new experiments Siavash did on face dataset?} \BL{Lotzi: we can include them, but we don't really have anything to compare with as the Berkeley paper doesn't do classification there, so our results are not directly comparable.} 

\end{compactitem}

\section{Related Work}
\label{sec:RelatedWork}

Few-shot or one-shot learning of classifiers has significant practical applications. Unfortunately, the few-shot learning model is not a good fit to the traditional training approaches of deep neural networks, which work best with large amounts of data. In recent years, significant research targeted approaches to allow deep neural networks to work in few-shot learning settings. One possibility is to perform transfer learning, but it was found that the accuracy decreases if the target task diverges from the trained task. One solution to mitigate this is through the use of an adversarial loss~\cite{luo2017label}.

A large class of approaches aim to enable few-shot learning by {\em meta-learning} - the general idea being that the meta-learning prepares the network to learn from the small amount of training data available in the few-shot learning setting. Note that meta-learning can be also used in other computer vision applications, such as fast adaptation for tracking in video~\cite{park2018meta}. The mechanisms through which meta-learning is implemented can be loosely classified in two groups. One class of approaches use a custom network architecture for encoding the information acquired during the meta-learning phase, for instance in fast weights~\cite{ba2016using}, neural plasticity values~\cite{miconi2018differentiable}, custom update rules~\cite{Metz-2018-LearningUnsupervisedLearning}, the state of temporal convolutions~\cite{Mishra-2018-SNAIL} or in the memory of an LSTM~\cite{ravi2016optimization}. The advantage of this approach is that it allows us to fine-tune the architecture for the efficient encoding of the meta-learning information. A disadvantage, however, is that it constrains the type of network architectures we can use; innovations in network architectures do not automatically transfer into the meta-learning approach. In a custom network architecture meta-learning model, the target learning phase is not the customary network learning, as it needs to take advantage of the custom encoding. 

A second, model-agnostic class of approaches aim to be usable for any differentiable network architecture. Examples of these algorithms are MAML~\cite{finn2017model} or Reptile~\cite{Nichol-2018-Reptile}, where the aim is to encode the meta-learning in the weights of the network, such that the network performs the target learning phase with efficient gradients. Approaches that customize the learning rates~\cite{Meier-2018-LearningRates} during meta-training can also be grouped in this class. For this type of approaches, the target learning phase uses the well-established learning algorithms that would be used if learning from scratch (albeit it might use specific hyperparameter settings, such as higher learning rates). 
We need to point out, however, that the meta-learning phase uses custom algorithms in these approaches as well (although they might use the standard learning algorithm in the inner loop, such as in the case of MAML). A recent work similar in spirit to ours is the CACTUs unsupervised meta-learning model described in~\cite{hsu2018unsupervised}. 

% This question is disingenous. There is a big difference between adjusting a hyperparameter vs being restricted to a particular architecture. Anyone using meta-learning will understand this. 
% \Question{Reviewer 3 comment:  I don’t understand. I thought the whole point was to not use custom algorithms, and you say so yourself. Why the contradiction?}

In this paper, we perform unsupervised meta-learning. Our approach generates tasks from unlabeled data which will help it to understand the structures of the relevant supervised tasks in the future. One should note that these relevant supervised tasks in the future do not have any intersection with the tasks which are used during the meta-learning. For instance, Wu~\etal perform unsupervised learning by recognizing a certain internal structure between dataset classes~\cite{wu2018unsupervised}. By learning this structure, the approach can be extended to semi-supervised learning. In addition, Pathak~\etal propose a method which learns object features in an interesting unsupervised way by detecting movement patterns of segmented objects~\cite{pathak2017learning}. These approaches are orthogonal to ours. We do not make assumptions that the unsupervised data shares classes with the target learning (in fact, we explicitly forbid it). Finally,~\cite{gupta2018unsupervised} define unsupervised meta-learning in reinforcement learning context. The authors study how to generate tasks with synthetic reward functions (without supervision) such that when the policy network is meta trained on them, they can learn real tasks with manually defined reward functions (with supervision) much more quickly and with fewer samples.

%\Question{
%Reviewer 3: -> The meta-testing performance is going to critically depend on the overlap between the imaginary task distribution and the testing distribution. Would help with clarity if this was mentioned early in the paper.
%}

% Assigning pseudo labels to data is proposed in~\cite{zhang2016understanding}. Zhang~\etal demonstrate that deep neural networks are powerful enough to shatter almost every dataset during training, however, the generalization is more complicated to understand. The assignment of random labels is just with the purpose of showing that neural networks can shatter any dataset even if the labels are assigned to wrong images.
    
% \SK{
%     \subsection{Unsupervised Learning}
    
%     In this section, we want to elaborate on the difference between our method and general unsupervised learning algorithms. 

%     \subsection{Pseudo Label Assignment}

% }
\section{The UMTRA algorithm}
\label{sec:Method}

\begin{notincluded}

\subsection{The few-shot classifier learning task}

As the UMTRA algorithm is based on generating few-shot classifier learning tasks, we need to start with a careful definition of this problem. Let us consider the objective of classifying samples, $\mathbf{x}$, drawn from a domain, $\mathbf{X}$, into classes, $\mathbf{y}_i \in Y = \{C_1, \ldots, C_N\}$. Without loss of generality, we  consider that the classes are encoded as one-hot vectors of dimensionality $N$. We are interested in learning a classifier $f_\theta$ that outputs a probability distribution over the classes. It is common to envision $f$ as a deep neural network parameterized by $\theta$, although this is not the only possible choice.

% (Siavash) Ask Dr. Boloni about this part.

We package a certain supervised learning task, $\mathcal{T}$, of type $(N, K)$, that is with $N$ classes of $K$ training samples each, as follows: We sample the training data of the form $(x_i,y_i)$, where $i=1\ldots N \times K$, $\mathbf{x}_i \in X$ and  $\mathbf{y}_i \in Y$. We assume that there are exactly $K$ samples for each possible $y_i$. In the recent meta-learning literature, it is often assumed that the task $\mathcal{T}$ has $K$ samples of each class for training {\em and} (separately), $K$ samples for validation $(x_j^v,y_j^v)$.

The choices above, including the equal split between the training and validation samples, and the symmetry of the distribution of the training samples and the fact that we call it an N-way K-shot classification, although we have N times 2K data if we include the validation data, are conventions to which we will adhere for easier comparison on the experimental results. 

%For this task, a conveniently defined loss function is the  cross-entropy loss over a particular dataset $\mathcal{D}$:
%\begin{equation}
%    \mathcal{L}(f_\phi) = -\sum_{x^{(j)},y^{(j)}\sim \mathcal{D}} y^{(j)}\log f_\phi(x^{(j)}) 
%\end{equation}
Tasks defined as above are of practical importance, because there are many domains in which the acquisition of the supervised training data is costly. For instance, it is possible that both the samples $(x_i, y_i)$ can only be collected indirectly, from human activity. Alternatively, the input $x_i$ can be provided by the algorithm, but a human needs to provide the $y_i$ part. Due to the cost of acquiring the training data, we are specially interested in solving problems with small $K$ sample values ({\em e.g} $K=5$ or even $K=1$). 

\subsection{Learning from scratch, transfer learning and supervised meta-learning}

With these definitions, a baseline {\em learning-from-scratch} approach would proceed as follows. We are given a task $\mathcal{T}$. We start from a randomly initialized classifier $f_{\theta_0}$, and update the value of $\theta$  through some kind of learning algorithm, until the loss is minimized on the training data of the task ($K$ samples of each $N$ classes) or after a certain number of iterations. We evaluate the performance by calculating the loss on the validation data of the task. Unfortunately, the lower the value of $K$, the lower the likelihood that a good classifier can be learned from scratch (that is, from a randomly initialized $\theta$). We need to start the learning process with a significant inductive bias which needs to be partially encoded in the classifier architecture and partially in its parameters $\theta$. 

A possible model is {\em transfer learning}, where the initialized $\theta$ was acquired by learning on a different problem. In practice, our expectation is that the cost of training $\theta$ has been already absorbed previously. This is the case, for instance of classifiers reusing ResNet or VGG networks trained on ImageNet. The way in which the transfer learning happened may be supervised or unsupervised. For transfer learning from models trained on ImageNet, this is, of course a supervised model. However, the general assumption here is that the learned $\theta$ in fact conveys more about the domain $X$ rather than the values $Y$.

{\em Meta-learning} models have a different objective from the transfer learning - the objective is that the learning of the task $\mathcal{T}$ starts with an architecture that is especially good at learning such tasks. When we talk about an architecture this involves not only the starting parameters $\theta_0$ but also learning rates, update rules, memory content and other rules that might be considered. A frequently encountered setup is the following: We assume that we have access to a collection of tasks $\mathcal{T}_1 \ldots \mathcal{T}_n$, drawn from a specific distribution of tasks. We meta-learn on these (supervised) tasks, and finally perform a task training on new task $\mathcal{T}$. Certain algorithms, such as MAML~\cite{finn2017model} use both the training and the validation data.

Whether it is worth-while to perform meta-learning before few-shot classifier learning according to this model depends on two questions:

\begin{compactitem}
    \item What is the {\em cost} of acquiring the dataset for meta-training tasks $\mathcal{T}_1 \ldots \mathcal{T}_n$? If the cost of acquiring this dataset is  higher than acquiring more training data for the target task $\mathcal{T}$, then we are better off by just acquiring training data for the task we are really interested in. 
    
    \item How {\em close} the meta-training tasks $\mathcal{T}_1 \ldots \mathcal{T}_n$ need to be to the target task $\mathcal{T}$? One of the most compelling use case of  few-shot learning is to perform classification in novel domains, where we either don't have enough samples $x$ or even humans might have difficulty assigning labels $y$. If we need to create many closely-related labeled tasks, this would restrict us to variations of well known domains. 
    
\end{compactitem}

\subsection{Unsupervised meta-learning for classification}

\end{notincluded}

% \SK{
\subsection{Preliminaries}
%As the UMTRA algorithm is based on generating few-shot classifier learning tasks, we need to start with a careful definition of this problem. 

We consider the task of classifying samples $\mathbf{x}$ drawn from a domain $\mathbf{X}$ into classes $\mathbf{y}_i \in Y = \{C_1, \ldots, C_N\}$. The classes are encoded as one-hot vectors of dimensionality $N$. We are interested in learning a classifier $f_\theta$ that outputs a probability distribution over the classes. It is common to envision $f$ as a deep neural network parameterized by $\theta$, although this is not the only possible choice.

% (Siavash) Ask Dr. Boloni about this part.

We package a certain supervised learning task, $\mathcal{T}$, of type $(N, K)$, that is with $N$ classes of $K$ training samples each, as follows. The training data will have the form $(x_i,y_i)$, where $i=1\ldots N \times K$, $\mathbf{x}_i \in X$ and  $\mathbf{y}_i \in Y$, with exactly $K$ samples for each value of $y_i$. In the recent meta-learning literature, it is often assumed that the task $\mathcal{T}$ has $K$ samples of each class for training {\em and} (separately), $K$ samples for validation $(x_j^v,y_j^v)$. 

% The choices above, including the equal split between the training and validation samples, and the symmetry of the distribution of the training samples and the fact that we call it an N-way K-shot classification, although we have N times 2K data if we include the validation data, are conventions to which we will adhere for easier comparison on the experimental results. 

In supervised meta-learning, we have access to a collection of tasks $\mathcal{T}_1 \ldots \mathcal{T}_n$ drawn from a specific distribution, with both supervised training and validation data.
The meta-learning phase uses this collection of tasks, while the target learning uses a new task $\mathcal{T}$ with supervised learning data but no validation data.

% have a different objective from the transfer learning - the objective is that the learning of the task $\mathcal{T}$ starts with an architecture that is especially good at learning such tasks. When we talk about an architecture this involves not only the starting parameters $\theta_0$ but also learning rates, update rules, memory content and other rules that might be considered. A frequently encountered setup is the following: 
% We assume that 
% Certain algorithms, such as MAML~\cite{finn2017model} use both the training and the validation data.

\subsection{Model}
%}

Unsupervised meta-learning retains the goal of meta-learning by preparing a learning system for the rapid learning of the target task $\mathcal{T}$. However, instead of the collection of tasks $\mathcal{T}_1 \ldots \mathcal{T}_n$ and their associated labeled training data, we only have an unlabeled dataset $\mathcal{U}=\{\ldots x_i \ldots\}$, with samples drawn from the same distribution as the target task. We assume that every element of this dataset is associated with a natural class $C_1 \ldots C_c$, $\forall x_i ~\exists j$ such that $x_i \in C_j$. We will assume that $N \ll c$, that is, the number of natural classes in the unsupervised dataset is much higher than the number of classes in the target task. These requirements are much easier to satisfy than the construction of the tasks for supervised meta-learning - for instance, simply stripping the labels from datasets such as Omniglot and Mini-ImageNet satisfies them.

The pseudo-code of the UMTRA algorithm is described in Algorithm~\ref{alg:UMTRA}. In the following, we describe the various parts of the algorithm in detail. In order to be able to run the UMTRA algorithm on unsupervised data, we need to create tasks $\mathcal{T}_i$ from the unsupervised data that can serve the same role as the meta-learning tasks serve in the full MAML algorithm. For such a task, we need to create both the training data $\mathcal{D}$ and the validation data $\mathcal{D}'$.

\AlgoDontDisplayBlockMarkers
%\SetAlgoNoEnd
\SetAlgoNoLine%
%\IncMargin{1em}
\begin{algorithm}[t]
\SetKwData{Left}{left}\SetKwData{This}{this}\SetKwData{Up}{up}
\SetKwFunction{Union}{Union}\SetKwFunction{FindCompress}{FindCompress}
\SetKwInOut{Require}{require}
%\SetKwInOut{Output}{output}
\Require{$N$: class-count, $N_\mathit{MB}$: meta-batch size, $N_\mathit{U}$: no. of updates}
\Require{$\mathcal{U}=\{\ldots x_i \ldots\}$ unlabeled dataset}
\Require{$\alpha$, $\beta$: step size hyperparameters}
\Require{$\mathcal{A}$: augmentation function}

%\BlankLine
randomly initialize $\theta$\;
\While{not done} {
\For{i in $1 \ldots N_\mathit{MB}$} {
    Sample $N$ data points $x_1 \ldots x_N$ from $\mathcal{U}$\;
    $\mathcal{T}_i \leftarrow \{x_1, \ldots x_N\}$\;
}
\ForEach{$\mathcal{T}_i$}{
Generate training set $\mathcal{D}_i=\{(x_1,1),\ldots,(x_N,N)\}$\;
$\theta_i' = \theta$\;
\For{j in $1 \ldots N_\mathit{U}$} {
Evaluate $\nabla_{\theta'_i}\mathcal{L}_{\mathcal{T}_i}(f_{\theta'_i})$\;
Compute adapted parameters with gradient descent: $\theta_i' = \theta_i' - \alpha \nabla_{\theta'_i}\mathcal{L}_{\mathcal{T}_i}(f_{\theta'_i}) $\;
}
Generate validation set for the meta-update $\mathcal{D}_i'=\{(\mathcal{A}(x_1),1),\ldots,(\mathcal{A}(x_N),N)\}$
}
Update $\theta \leftarrow \theta - \beta\nabla_\theta \sum_{\mathcal{T}_i} \mathcal{L}_{\mathcal{T}_i}(f_{\theta'_i})$ using each $\mathcal{D}'_i$\;
% (Siavash I commented this part out)
% and $\mathcal{L}_{\mathcal{T}_i}$\;
}
\caption{Unsupervised Meta-learning with Tasks constructed by Random sampling and Augmentation (UMTRA)}\label{alg:UMTRA}
\end{algorithm}
%\DecMargin{1em}

\medskip

\noindent{\bf Creating the training data:} In the original form of the MAML algorithm, the training data of the task $\mathcal{T}$ must have the form $(x,y)$, and we need $N \times K$ of them. The exact labels used during the meta-training step are not relevant, as they are discarded during the meta-training phase. They can be thus replaced with artificial labels, by setting them $y \in \{1, ... N\}$. It is however, important that the labels maintain class distinctions: if two data points have the same label, they should 
% be in the same class, while if they have different labels, they should be in different classes. 
also have the same artificial labels, while if they have different labels, they should have different artificial labels.

The first difference between UMTRA and MAML is that during the meta-training phases, we always perform one-shot learning, with $K=1$. Note that during the target learning phase we can still set values of $K$ different from 1. The training data is created as the set $\mathcal{D}_i = \{(x_1,1), \ldots (x_N, N)\}$, with $x_{i}$ {\em sampled randomly} from $\mathcal{U}$. 

Let us see how this training data construction satisfy the class distinction conditions. The first condition is satisfied because there is only one sample for each label. The second condition is satisfied statistically by the fact that $N \ll c$, where $c$ is the total number of classes in the dataset. If the number of samples is significantly smaller than the number of classes, it is likely that all the samples will be drawn from different classes. If we assume that the samples are equally distributed among the classes (e.g. $m$ samples for each class), the probability that all samples are in a different class is equal to 
\begin{equation}
P = \frac{(c \cdot m) \cdot ((c - 1) \cdot m) ... ((c - N + 1) \cdot m)}{(c \cdot m) \cdot (c \cdot m - 1) ... (c \cdot m - N + 1)} = \frac{c! \cdot m^N \cdot (c \cdot m - N)!}{(c - N)! \cdot (c \cdot m)!}
\end{equation}

 To illustrate this, the probability for 5-way classification on the Omniglot dataset used with each of the 1200 characters is a separate class ($c=1200$, $N=5$) is 99.21\%. For Mini-ImageNet ($c=64$), the probability is 85.23\%, while for the full ImageNet it would be about 99\%.

% {\color{green} Omniglot has 30 different alphabets in its training division, but we consider each character as a different class which makes $c=1200, N=5$. On the other hand, we use Mini-Imagenet which has $c=100, N=5$ but we use 64 classes for training which makes it $c=64, N=5$}

\medskip
\noindent{\bf Creating the validation data:} For the MAML approach, the validation data of the meta-training tasks is actually training data in the outer loop. It is thus required that we create a validation dataset $\mathcal{D}_i' = \{(x_1',1), \ldots (x_N', N)\}$ for each task $\mathcal{T}_i$. Thus we need to create appropriate validation data for the synthetic task. A minimum requirement for the validation data is to be correctly labeled in the given context. This means that the synthetic numerical label should map in both cases to the same class in the unlabeled dataset: $\nexists~C$ such that $x_i, x_i' \in C$.

In the original MAML model, these $x_i'$ values are labeled examples part of the supervised dataset. In our case, picking such $x_i'$ values is non-trivial, as we don't have access to the actual class. Instead, we propose to {\em create} such a sample by augmenting the sample used in the training data using an {\em augmentation function} $x_i' = \mathcal{A}(x_i)$ which is a hyperparameter of the UMTRA algorithm. A requirement towards the augmentation function is to maintain class membership $x \in C \Rightarrow \mathcal{A}(x) \in C$. We should aim to construct the augmentation function to verify this property for the given dataset $\mathcal{U}$, based on what we know about the domain described by the dataset. However, as we do not have access to the classes, such a verification is not practically possible on a concrete dataset.

% A domain independent choice for the augmentation function is the identity function $\mathcal{A} = \mathbbm{1}$, that is, repeating the training data as validation data. While this would be an unsound practice for ``validation" data understood in the usual sense, we need to point out that the validation data of the MAML inner loop is actually used as a training data in the outer loop. Thus the $\mathcal{A} = \mathbbm{1}$ setting simply makes us use the same data in the (very different) updates in lines 12 and lines 16 of the Algorithm~\ref{alg:UMTRA}. Unfortunately, as we explain in section~\ref{sub:theory}, this is not going to work due to biased estimation of the generalization error.

% The advantage of the $\mathcal{A} = \mathbbm{1}$ is that it is theoretically applicable to any dataset - be that images or videos, and it does not depend on the domain. \SK{Based on our theory part, this augmentation function should not work well, so maybe we want to rephrase this paragraph} The disadvantage is that this setting can lead to a sort of ``meta-overfitting''. In traditional overfitting, we are overfitting to the training data -- this is not an issue here, because we do not have the training data in the meta-learning phase. Instead, the danger is that the network will learn that the tasks are defined very narrowly. In some ways, the augmentation function is a way to complicate the tasks and force the network to learn better features.
% In some ways, the augmentation function is a way to convey to the network the approximate size of the classes.

\begin{figure}
    \centering
    \includegraphics[height=0.22\columnwidth]{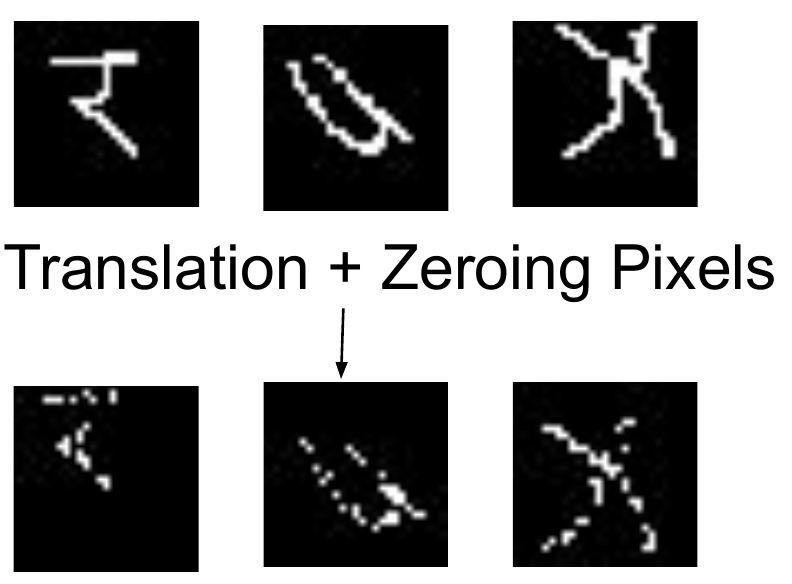}
    \hspace{10pt}
    \vline
    \hspace{15pt}
    \includegraphics[height=0.22\columnwidth]{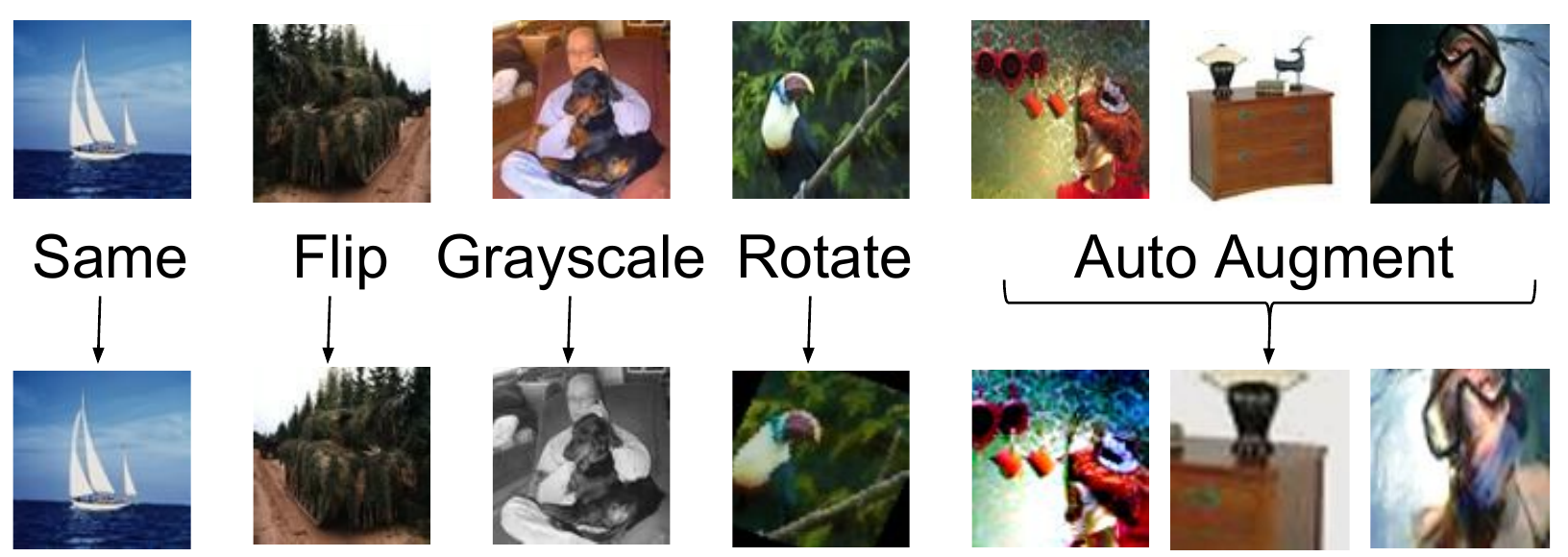}

    \caption{
    Augmentation techniques on Omniglot (left) and Mini-Imagenet (right). 
    Top row: Original images in training data. Bottom: augmented images for the validation set, transformed with an augmentation function $\mathcal{A}$. Auto Augment~\cite{cubuk2018autoaugment} applies augmentations from a learned policy based on combinations of translation, rotation, or shearing.}
    \label{fig:AugmentedImages}
\end{figure}

Another choice for the augmentation function $\mathcal{A}$ is to apply some kind of domain-specific change to the images or videos. Examples of these include setting some of the pixel values to zero in the image (Figure~\ref{fig:AugmentedImages}, left), or translating the pixels of the training image by some amount (eg. between -6 and 6). 

The overall process of generating the training data from the unlabeled dataset in UMTRA and the differences from the supervised MAML approach is illustrated in Figure \ref{fig:story}.

% \begin{figure}
% \hspace{2cm} Supervised MAML \hspace{5cm} UMTRA\\
% \begin{center}
% \includegraphics[width=0.49\columnwidth]{images/SupervisedMetaTraining.pdf}
% \includegraphics[width=0.49\columnwidth]{images/UnsupervisedMetaTraining.pdf}
% \end{center}
% \caption{The process of creation of the training and validation data of the meta-training task $\mathcal{T}$. (top) Supervised MAML: We start from a dataset where the samples are labeled with their class. The training data is created by sampling $N$ distinct classes $C_{L_i}$, and choosing a random sample $x_i$ from each. The validation data is created by choosing a different sample $x_i'$ from the same class. (bottom) UMTRA: We start from a dataset of unlabeled data. The training data is created by randomly choosing N samples $x_i$ from the dataset (and hoping that they are of different classes). The validation data is created by applying the augmentation function $\mathcal{A}$ to each sample from the training data. For both MAML and UMTRA, artificial temporary labels $1, 2 \ldots N$ are used. 
% }
% \label{fig:story}
% \end{figure}

\subsection{Some theoretical considerations}
\label{sub:theory}

While a full formal model of the learning ability of the UMTRA algorithm is beyond the scope of this paper, we can investigate some aspects of its behavior that shed light into why the algorithm is working, and why augmentation improves its performance. Let us denote our network with a parameterized function $f_{\theta}$. As we want to learn a few-shot classification task, $\mathcal{T}$ we are searching for the corresponding  function $f_{\mathcal{T}}$, to which we do not have access. To learn this function, we use the training dataset, $D_{\mathcal{T}} = {\{(x_{i}, y_{i})\}}_{i=1}^{n \times k}$. For this particular task, we update our parameters (to $\theta'$) to fit this task's training dataset. In other words, we want $f_{\theta'}$ to be a good approximation of $f_{\mathcal{T}}$. 
Finding $\theta'$ such that, $\displaystyle\theta' = \argmin_{\theta} \sum_{(x_{i}, y_{i}) \in D_{\mathcal{T}}} {\mathcal{L}(y_{i}, f_{\theta}(x_{i}))}$ is ill-defined because there are more than one solution for it. In meta-learning, we search for the $\theta'$ value that gives us the minimum generalization error, the measure of how accurately an algorithm is able to predict outcome values for unseen data~\cite{abu2012learning}. We can estimate the generalization error based on sampled data points from the same task. Without loss of generality, let us consider a sampled data point $(x_{0}, y_{0})$. We can estimate generalization error on this point as $\mathcal{L}(y_{0}, f_{\theta'}(x_{0}))$. In case of mean squared error, and by accepting irreducible error $\epsilon \sim \mathcal{N}(0, \sigma)$, we can decompose the expected generalization error as follows~\cite{james2013introduction,friedman2001elements}:
\begin{equation}
    E\left[\mathcal{L}(y_{0}, f_{\theta'}(x_{0}))\right] = \big({E[{f_{\theta'}}(x_0)] - f_{\mathcal{T}}(x_{0})}\big)^2 + E\left[\left({f_{\theta'}(x_0)}\right)^2\right] - {E\left[f_{\theta'}(x_0)\right]}^2 + \sigma^2
    \label{eq:2}
\end{equation}
In this equation, when $(x_{0}, y_{0}) \notin D_\mathcal{T}$ we have $E[\left({f_{\theta'}(x_0)}\right)^2] - {E[f_{\theta'}(x_0)]}^2 = 0$, which means that the estimation of the generalization error on these samples will be as unbiased as possible (only biased by $\sigma^2$). On the other hand, if $(x_{0}, y_{0}) \in D_\mathcal{T}$, the estimation of the error is going to be highly biased. We conjecture that similar results will be observed for other loss functions as well with the estimate of the loss function being more biased if the samples are from the training data rather than outside it. As in the outer loop of MAML estimates the generalization error on a validation set for each task in a batch of tasks, it is important to keep the validation set separate from the training set, as this estimate will be eventually applied to the starter network. 

%For loss functions other than mean squared error, it is sill true that the estimation will be biased if we choose $(x_{0}, y_{0}) \in D_{\mathcal{T}}$.
% This means by sampling a validation set separate from train set for each task we can have a good estimation of the generalization error. This is similar to the outer loop error of the MAML algorithm. The outer loop error is the estimation of the generalization error on a validation set for each task in a batch of tasks. 
% \SK{We can show that this term is related to the complexity of the model and that is why regularization helps.}
In contrast, if we pick our validation set as points in $D_\mathcal{T}$, our algorithm is going to learn to minimize a biased estimation of the generalization error. Our experiments also show that if we choose the same data for train and test ($\mathcal{A}(x) = x$), we will end up with an accuracy almost the same as training from scratch. UMTRA, however, tries to improve the estimation of generalization error with augmentation techniques. Our experiments show that by applying UMTRA with good choice of function for augmentation, we can achieve comparable results with supervised meta-learning algorithms. In our supplementary material, we show that UMTRA is able to adapt very quickly with just few iterations to a new task. Last but not least, in comparison with CACTUs algorithm which applies advanced clustering algorithms such as DeepCluster~\cite{caron2018deep}, ACAI~\cite{berthelot2018understanding}, and BiGAN~\cite{donahue2016adversarial} to generate train and validation set for each task, our method does not require clustering. 

% We also visualize the t-SNE of last hidden layer which shows that UMTRA is able to learn feature spaces which gives good generality.

% The main question is what is the effect of this term on $(\mathcal{A}(x_{0}), y_{0})$ for $(x_{0}, y_{0}) \in D_{\mathcal{T}}$. The answer to this question is in the augmentation function. If the augmentation function is able to generate new samples from $f_{task}$ which are separated from the given input, then the loss would be the same as supervised MAML. We do not have such an augmentation technique, however, there are good augmentation techniques which describe new instances that will result in improving the accuracy of the model\SK{cite some papers about augmentation here}. If augmentation function is able to create such instances, even though the estimation will be biased, but the bias is lower and the signal for meta-learning has valuable information. Our experiments, show that if we meet such conditions, we can achieve very good results on the target task.

\section{Experiments}
\label{sec:Experiments}

\subsection{UMTRA on the Omniglot dataset}

Omniglot~\cite{lake2011one} is a dataset of handwritten characters frequently used to compare few-shot learning algorithms. It comprises 1623 characters from 50 different alphabets. Every character in Omniglot has 20 different instances each was written by a different person. To allow comparisons with other published results, in our experiments we follow the experimental protocol described in \cite{santoro2016meta}: 1200 characters were used for training, 100 characters were used for validation and 323 characters were used for testing. 

UMTRA, like the supervised MAML algorithm, is model-agnostic, that is, it does not impose conditions on the actual network architecture used in the learning. This does not, of course, mean that the algorithm performs identically for every network structure and dataset. In order to separate the performance of the architecture and the meta-learner, we run our experiments using an architecture originally proposed in~\cite{vinyals2016matching}. This classifier uses four 3 x 3 convolutional modules with 64 filters each, followed by batch normalization~\cite{ioffe2015batch}, a ReLU nonlinearity and 2 x 2 max-pooling. On the resulting feature embedding, the classifier is implemented as a fully connected layer followed by a softmax layer. 

UMTRA has a relatively large hyperparameter space that includes the augmentation function. As pointed out in a recent study involving performance comparisons in semi-supervised systems~\cite{oliver2018realistic}, excessive tuning of hyperparameters can easily lead to an overestimation of the performance of an approach compared to simpler approaches. Thus, for the comparison in the remainder of this paper, we keep a relatively small budget for hyperparameter search: beyond basic sanity checks, we only tested 5-10 hyperparameter combinations per dataset, without specializing them to the N or K parameters of the target task. Table~\ref{tab:AugmentationCompare}, left, shows several choices for the augmentation function for the 5-way one-shot classification on Omniglot. Based on this table, in comparing with other approaches, we use an augmentation function consisting of randomly zeroed pixels and random shift.

% As a note, based on the spread of the accuracies in Table~\ref{tab:AugmentationCompare}, it is very likely that other choices of hyperparameters, possibly specialized to the $N$ and $K$ values may offer higher accuracy values. Depending on the needs of an application and the computational budget available, in practice it may be worthwhile to perform such a hyperparameter search step. This however, is beyond the scope of this paper. 

\begin{table}
    \caption{The influence of augmentation function on the accuracy of UMTRA for 5-way one-shot classification on the (Left: Omniglot dataset, Right: Mini-Imagenet dataset). For all cases, we use meta-batch size $N_\mathit{MB}=4$ and number of updates $N_U=5$, except the ones with best hyperparameters.}

    \centering
    {\footnotesize
    %\begin{tabular}{|p{5cm}|p{1cm}|}
    \begin{tabular}{ll}
     \begin{tabular}{p{4cm}p{1.2cm}}
         \hline 
         {\bf Augmentation Function $\mathcal{A}$} & 
        %  {\bf Meta-batch size $N_\mathit{MB}$} &
        % {\bf No. of\newline updates $N_U$} &
         {\bf Accuracy} \\
         \hline
         {\em Training from scratch} &
        %   N/A & 
        %   N/A &
         52.50 \\
%         \hline
         $\mathcal{A} = \mathbbm{1}$ & 
        %   1 & 
        %   1 & 
         52.93 \\ 
%         \hline
         $\mathcal{A}$ = randomly zeroed pixels & 
        %   1 & 
        %   1 &
         56.23 \\
%         \hline
         $\mathcal{A}$ = randomly zeroed pixels (with best hyperparameters) & 
        %   1 & 
        %   1 &
         67.00 \\
        %  \hline
        % $\mathcal{A}$ = 
        %  randomly zeroed pixels 
        %  \newline $~~~~~\quad$
        % + random shift & 
        % %  1 & 1 & 
        % 67.08 \\
%         \hline
         $\mathcal{A}$ = randomly zeroed pixels 
          \newline $~~~~~\quad$
        + random shift (with best \newline $~~~~~\quad$ hyperparameters) & 
        %   25 & 10 & 
         \textbf{83.80} \\
%         \hline
         {\em Supervised MAML} & 
         98.7 \\
         \hline
    \end{tabular}
    &
    \begin{tabular}{p{4cm}p{1.2cm}}
         \hline 
         {\bf Augmentation Function $\mathcal{A}$}  &  {\bf Accuracy} \\
         \hline
         {\em Training from scratch} &  24.17 \\
%         \hline
         $\mathcal{A} = \mathbbm{1}$ & 26.49 \\
%         \hline
         $\mathcal{A}$ = Shift + random flip & 30.16 \\
%         \hline
         $\mathcal{A}$ = Shift + random flip + randomly change to grayscale & 32.80 \\
%         \hline
         $\mathcal{A}$ = Shift + random flip + random rotation + color distortions & 35.09 \\
 %        \hline
         $\mathcal{A}$ = Auto Augment~\cite{cubuk2018autoaugment} & \textbf{39.93} \\
 %        \hline
         {\em Supervised MAML} &  46.81 \\
         \hline
    \end{tabular}   
         
    \end{tabular}
    
    }
    \label{tab:AugmentationCompare}
\end{table}

% \begin{table}
%     \centering
%     {\footnotesize
    
%     }
%     \vspace{1mm}
%     \caption{Exploration of several choices for the augmentation function hyperparameter on the Mini-Imagenet classification. For all cases, we use meta-batch size $N_\mathit{MB}=4$ and number of updates $N_U=5$.}
%     \label{tab:AugmentationCompareMiniImageNet}
% \end{table}

In our experiments, we realized two of the most important hyperparameters in meta-learning are meta-batch size, $N_{MB}$, and number of updates, $N_{U}$.
In table \ref{tab:hyperparameter-compare-omniglot}, we study the effects of these hyperparameters on the accuracy of the network for the randomly zeroed pixels and random shift augmentation. Based on this experiment, we decide to fix the meta-batch size to 25 and number of updates to 1. 
\begin{table}
    \caption{The effect of hyperparameters meta-batch size, $N_{MB}$, and number of updates, $N_{U}$ on accuracy. Omniglot 5-way one shot.}
    \centering
    {\footnotesize
        \begin{tabular}{c p{0.6cm}p{0.6cm}p{0.6cm}p{0.6cm}p{0.6cm}p{0.6cm}}
        \hline
        \diagbox{\scriptsize{$\#$ Updates}}{\scriptsize{$N_\mathit{MB}$}} & 1 & 2 & 4 & 8 & 16 & 25 \\
        \hline
        1 & 67.08 & 79.04 & 80.72 & 81.60 & 82.72 & {\bf 83.80} \\
        %\hline
        5 & 76.08 & 76.68 & 77.20 & 79.56 & 81.12 & 83.32 \\
        %\hline
        10 & 79.20 & 79.24 & 80.92 & 80.68 & 83.52 & 83.26\\
        \hline
        \end{tabular}
    }
    \label{tab:hyperparameter-compare-omniglot}
\end{table}

In order to find out the relationship between the level of the augmentation and accuracy, we apply different levels of augmentation on images. If the generated samples are different from current observation but within the same class manifold, UMTRA performs well. The results of this experiment are shown in table~\ref{tab:distortion_effects}.

\begin{table}
    \caption{The effect of the augmentation level on UMTRA's accuracy on the Omniglot dataset. In all of the experiments we use random pixel zeroing with meta-batch size $N_\mathit{MB}=25$ and number of updates $N_U=1$. }
    \centering
    {\footnotesize
    \begin{tabular}{cccccccc}
         \hline 
         {\bf Translation Range (Pixels)}  & 0 & 0-3 & 3-6 & 0-6 & 6-9 & 9-12 & 0-9  \\
         \hline
          {\bf Accuracy~\%} & 67.0 & 82.8 & 80.4 & {\bf 83.8} & 79.8 & 77 & 80.4 \\
         \hline
         
    \end{tabular}
    }
    \label{tab:distortion_effects}
\end{table}

The second consideration is what sort of baseline we should use when evaluating our approach on a few-shot learning task? Clearly, supervised meta-learning approaches such as an original MAML~\cite{finn2017model} are expected to outperform our approach, as they use a labeled training set. A simple baseline is to use the same network architecture being trained from scratch with only the final few-shot labeled set. If our algorithm takes advantage of the unsupervised training set $\mathcal{U}$, as expected, it should outperform this baseline. 

A more competitive comparison can be made against networks that are first trained to obtain a favorable embedding using unsupervised learning on $\mathcal{U}$, with the resulting embedding used on the few-shot learning task. These baselines are not meta-learning approaches, however, we can train them with the same target task training set as UMTRA. Similar to~\cite{hsu2018unsupervised}, we compare the following unsupervised pre-training approaches: ACAI~\cite{berthelot2018understanding}, BiGAN~\cite{donahue2016adversarial}, DeepCluster~\cite{caron2018deep} and InfoGAN~\cite{chen2016infogan}. These up-to-date approaches cover a wide range of the recent advances in the area of unsupervised feature learning. Finally, we also compare against the CACTUs unsupervised meta-learning algorithm proposed in the \cite{hsu2018unsupervised}, combined with MAML and ProtoNets~\cite{snell2017prototypical}. As a note, another unsupervised meta-learning approach related to UMTRA and CACTUs is AAL~\cite{antoniou2019assume}. However, as \cite{antoniou2019assume} doesn't compare against stock MAML, the results are not directly comparable. 

Table~\ref{tab:ResultsOmniglot}, columns three to six, shows the results of the experiments. For the UMTRA approach we trained for 6000 meta-iterations for the 5-way, and 36,000 meta-iterations for the 20-way classifications. Our approach, with the proposed hyperparameter settings outperforms, with large margins, training from scratch and the approaches based on unsupervised representation learning. UMTRA also outperforms, with a smaller margin, the CACTUs approach on all metrics, and in combination with both MAML and ProtoNets. 

As expected, the supervised meta-learning baselines perform better than UMTRA. To put this value in perspective, we need to take into consideration the vast difference in the number of labels needed for these approaches. In 5-way one-shot classification, UMTRA obtains a 83.80\% accuracy with only 5 labels, while supervised MAML obtains 94.46\% but requires 24005 labels. For 5-way 5-shot classification UMTRA obtains a 95.43\% accuracy with only 25 labels, while supervised MAML obtains 98.83\% with 24025. 

%UMTRA-mod  with additional FC layer 2000 iterations (ours).

\begin{table}
    \centering
    \caption{Accuracy in \% of N-way K-shot (N,K) learning methods on the Omniglot and Mini-Imagenet datasets. The ACAI~/~DC label means ACAI Clustering on Omniglot and DeepCluster on Mini-Imagenet. The source of non-UMTRA values is~\cite{hsu2018unsupervised}. }    
    {\footnotesize
    \begin{tabular}{p{2.9cm}p{1.4cm}p{0.6cm}p{0.6cm}p{0.6cm}p{0.6cm}|p{0.6cm}p{0.6cm}p{0.6cm}p{0.6cm}}
        \hline
         &  &\multicolumn{4}{c}{\bf Omniglot} & \multicolumn{4}{c}{\bf Mini-Imagenet} \\
        \hline
        {\bf Algorithm (N, K)} & {\bf Clustering} & {\bf (5,1)} & {\bf (5,5)} & \bf{(20,1)} & \bf{(20,5)} & {\bf (5,1)} & {\bf (5,5)} & \bf{(5,20)} & \bf{(5,50)}\\
        \hline
        {\em Training from scratch} & N/A & 52.50 & 74.78 & 24.91 & 47.62 &
        % 24.17 
         27.59 &
        %  38.48
        38.48
        %  &  51.53 
        &
        51.53
         & 
        %  59.63
        59.63\\
        \hline
        $k_{nn}$-nearest neighbors & BiGAN & 49.55 & 68.06  & 27.37 & 46.70 & 25.56 & 31.10 & 37.31 & 43.60\\
        linear classifier & BiGAN & 48.28 & 68.72  & 27.80 & 45.82 & 27.08 & 33.91 & 44.00 & 50.41\\
        MLP with dropout & BiGAN & 40.54 & 62.56  & 19.92 & 40.71 & 22.91 & 29.06 & 40.06 & 48.36\\
        cluster matching & BiGAN & 43.96 & 58.62  & 21.54 & 31.06 & 24.63 & 29.49 & 33.89 & 36.13 \\
        CACTUs-MAML & BiGAN  & 58.18 & 78.66  & 35.56 & 58.62 & 36.24 & 51.28 & 61.33 & 66.91 \\
        CACTUs-ProtoNets & BiGAN & 54.74 & 71.69  & 33.40 &  50.62 & 36.62 & 50.16 & 59.56 & 63.27\\
        % Siavash, please leave it like this
        % &&&&\\
        $k_{nn}$-nearest neighbors & ACAI~/~DC & 57.46 & 81.16 & 39.73 & 66.38 & 28.90 & 42.25 & 56.44 & 63.90\\
        linear classifier & ACAI~/~DC & 61.08 & 81.82  &  43.20 & 66.33 & 29.44 & 39.79 & 56.19 & 65.28\\
        MLP with dropout & ACAI~/~DC & 51.95 & 77.20  &  30.65 & 58.62 & 29.03 & 39.67 & 52.71 & 60.95 \\
        cluster matching & ACAI~/~DC & 54.94 & 71.09  & 32.19 &  45.93 & 22.20 & 23.50 & 24.97 & 26.87 \\
        %\hline
        CACTUs-MAML & ACAI~/~DC & 68.84 &  87.78  & 48.09 & 73.36 & 39.90 & {\bf 53.97} & {\bf 63.84} & {\bf 69.64} \\
        CACTUs-ProtoNets & ACAI~/~DC & 68.12 & 83.58  & 47.75 & 66.27 & 39.18 & 53.36 & 61.54 & 63.55 \\
        % Siavash, please leave it like this
        % &&&&\\
        \hline
        UMTRA (ours) & N/A & \textbf{83.80} & \textbf{95.43} & \textbf{74.25} & \textbf{92.12} & \textbf{39.93} & 50.73 & 61.11 & 67.15\\
        % UMTRA (ours) altern. arch. & \textbf{80.16} & \textbf{90.68} & \textbf{50.89} & \textbf{77.93} \\
        \hline
        % AAL-MAML+++CHVW~\footnote{This method is bulit on top of MAML++, not stock MAML, so it is not directly comparable.}&  88.40 & 97.96 & 70.21 & 88.32 \\
        {\em MAML (Supervised)} & N/A & 94.46 & 98.83 &  84.60 &  96.29 & 46.81 & 62.13 & 71.03 & 75.54\\
        {\em ProtoNets (Supervised)} & N/A & 98.35 & 99.58  & 95.31 & 98.81 & 46.56 & 62.29 & 70.05 & 72.04 \\
        \hline
        \end{tabular}
        
    }
    \label{tab:ResultsOmniglot}
\end{table}

\begin{notincluded}

\subsubsection{Analyze the effect of distortion on spanning the manifold of the classes}

In this section, we apply different distortions on the input and try to figure out what would be the effect of these distortions on the data. Generally, if we are able to augment samples which are different from current observation but within the same class manifold, we can get good result on UMTRA. We have distorted the images with different percentage of pixel zeroing and translation and show the results in table~\ref{tab:distortion_effects}.

\begin{table}
    \centering
    {\footnotesize
    \begin{tabular}{|c|c|}
         \hline 
         {\bf Distortion Percentage~\%}  &  {\bf Accuracy~\%} \\
         \hline
         25.00 &  61.4 \\
         \hline
         26.78 &  71.0 \\
         \hline
         30.35 &  81.2 \\
         \hline
         80 & 86.0  \\
%         \hline
%         100 &   \\
         \hline
         
    \end{tabular}
    }
    \vspace{1mm}
    \caption{Effect of distortion on Accuracy of UMTRA. With good enough distortion which still stays in the same class, however, generates new instances, the algorithm is able to perform very well.}
    \label{tab:distortion_effects}
\end{table}

To compare generalization of training from scratch, UMTRA and supervised MAML, we visualize the activations of the last hidden layer of the network on Omniglot dataset by t-SNE. We compare all of the methods on the same target training task which is constructed by sampling five characters from test data and selecting one image from each character class randomly. Each character has 20 different instances. Figure~\ref{fig:omniglot_raw_pixel_tsne} shows the t-SNE visualization of raw pixel values of these 100 images. Instances which are sampled for the one-shot learning task are connected to each other by dotted lines. Figure~\ref{fig:omniglot_tsne} shows the visualization of the last hidden layer activations for the same task. UMTRA as well as MAML can adapt quickly to a feature space which has a better generalization than training from scratch.

\begin{figure}[ht]
    \centering
    \includegraphics[width=0.5\columnwidth]{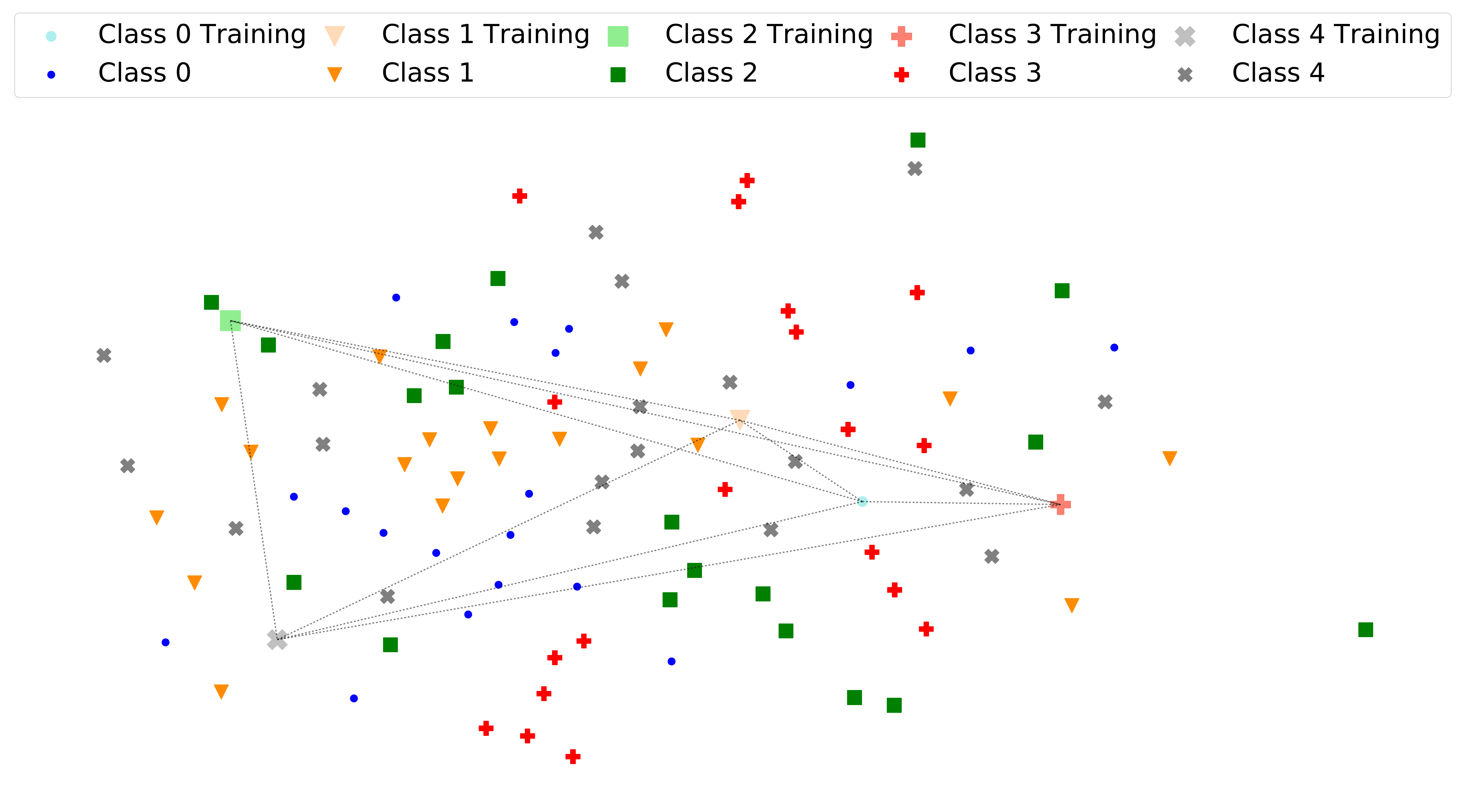}
    \caption{
        t-SNE on the Omniglot raw pixel values.
        % Training instances are denoted by larger and lighter symbol and are connected to each other by dotted lines
        % The target task contains 5 different classes. The class instances are visualized with different colors and shapes. The five instances used for target task training are connected to each other with dotted lines.
    }
    \label{fig:omniglot_raw_pixel_tsne}
\end{figure}

\begin{figure}
    \centering
    
    Training from Scratch
    
    \includegraphics[width=0.49\columnwidth]{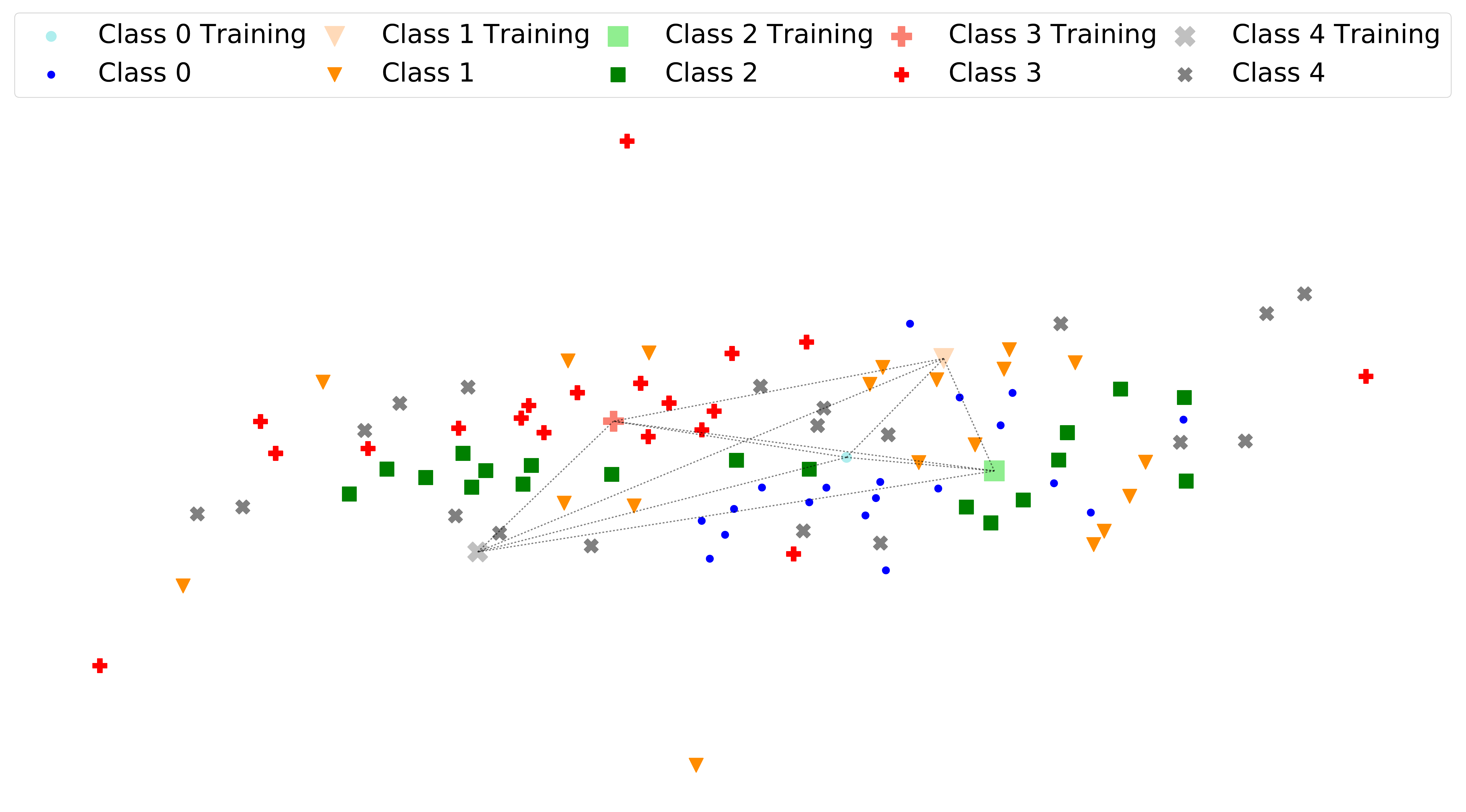}
    \includegraphics[width=0.49\columnwidth]{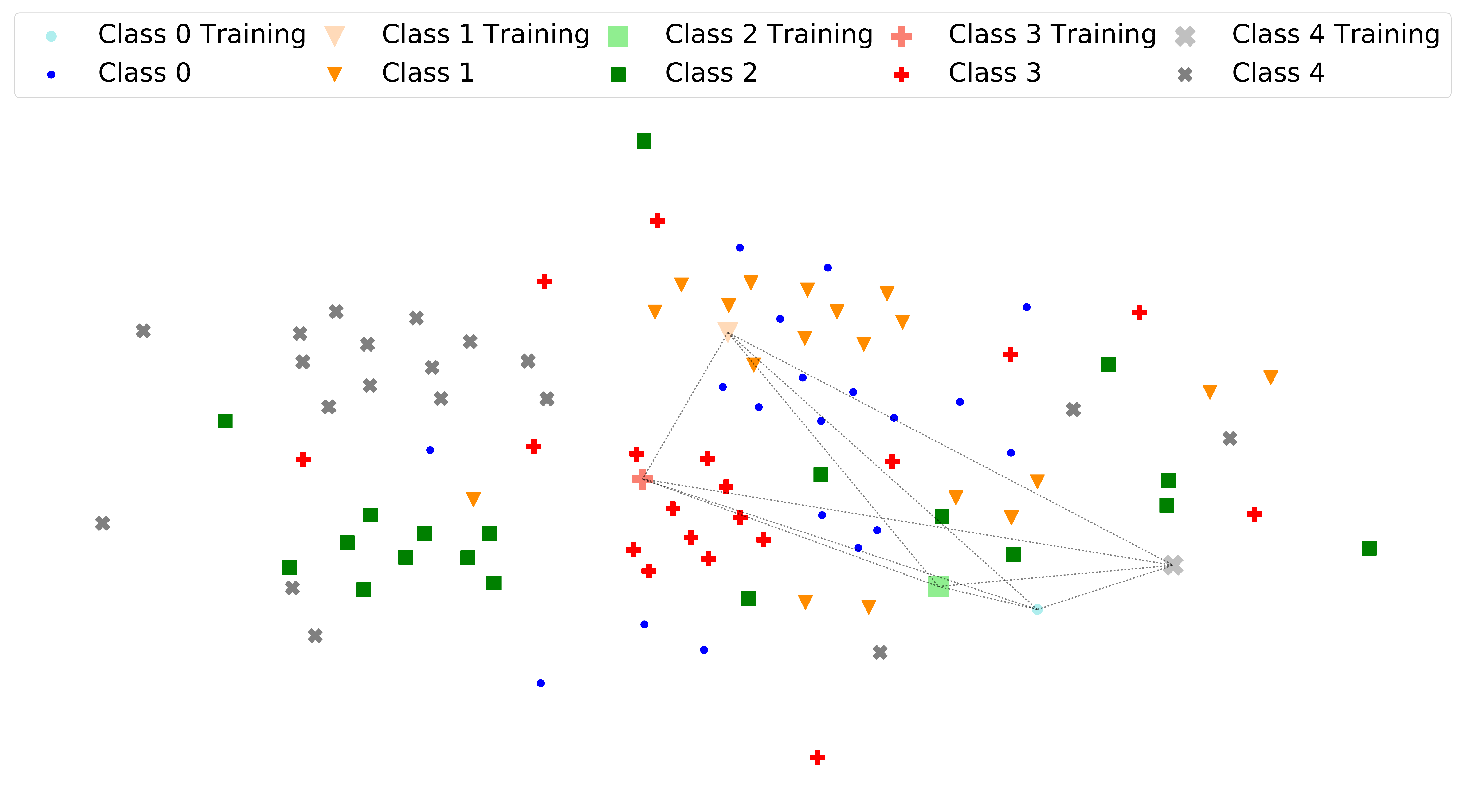}
    
    UMTRA
    
    \includegraphics[width=0.49\columnwidth]{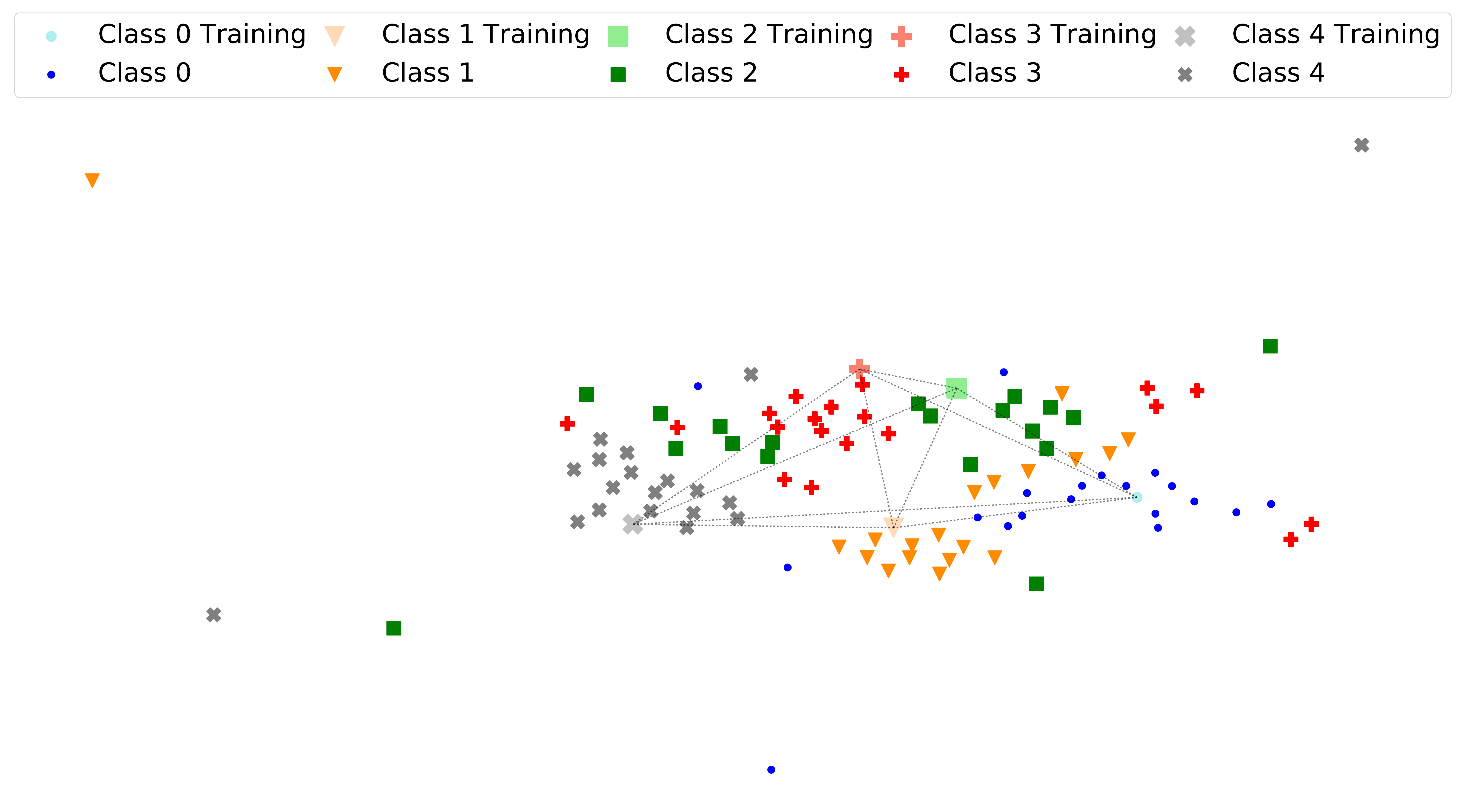}
    \includegraphics[width=0.49\columnwidth]{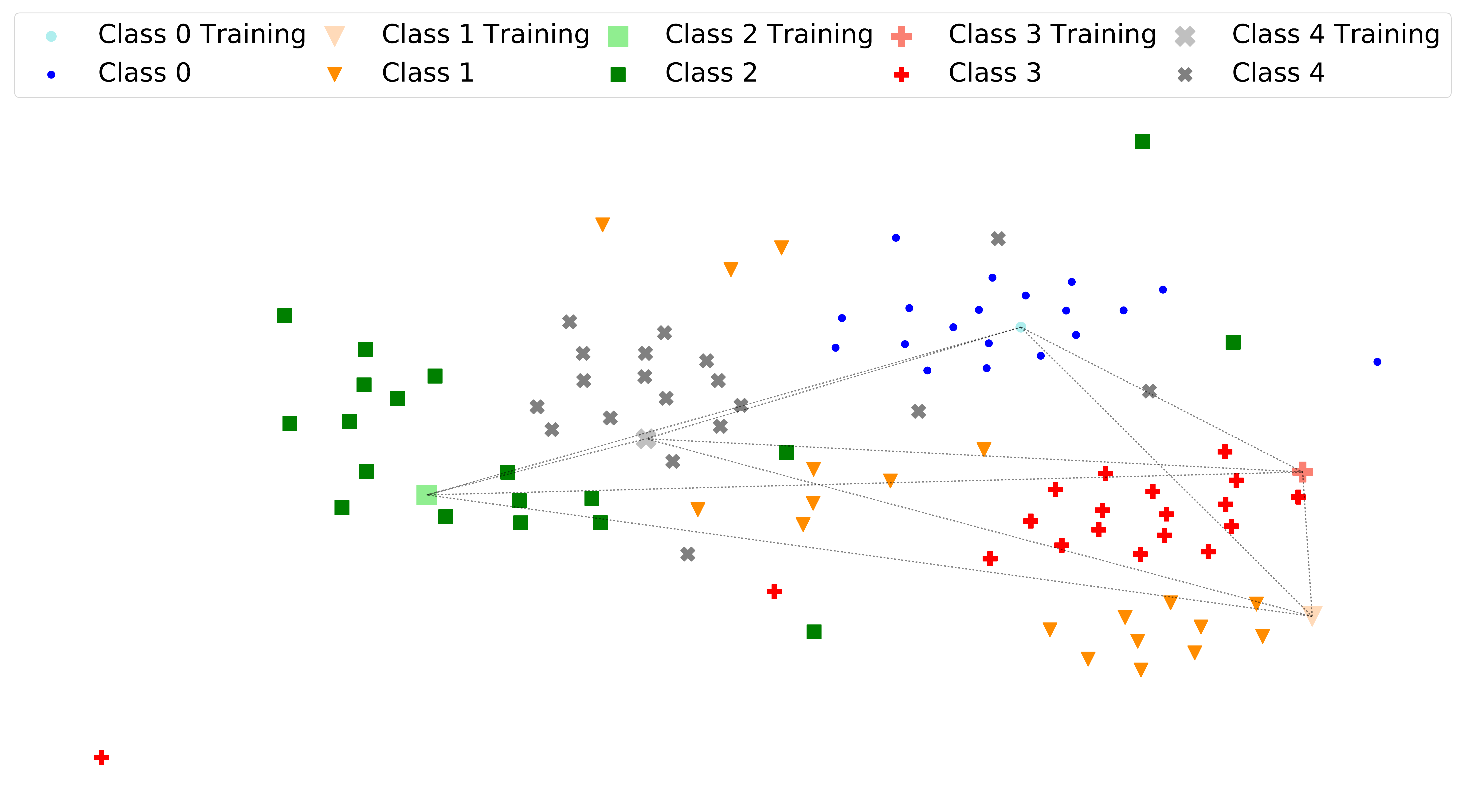}
    
    MAML
    
    \includegraphics[width=0.49\columnwidth]{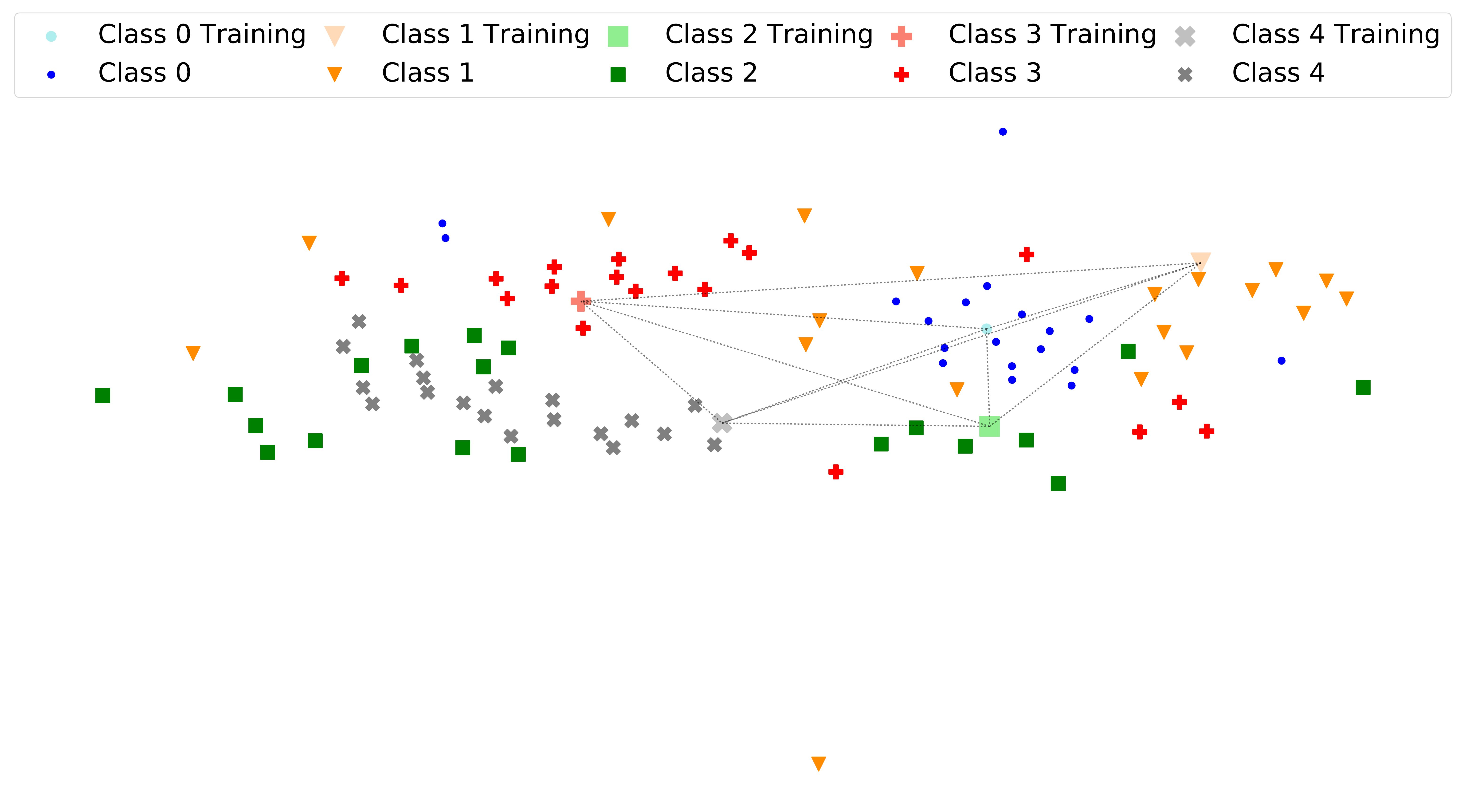}
    \includegraphics[width=0.49\columnwidth]{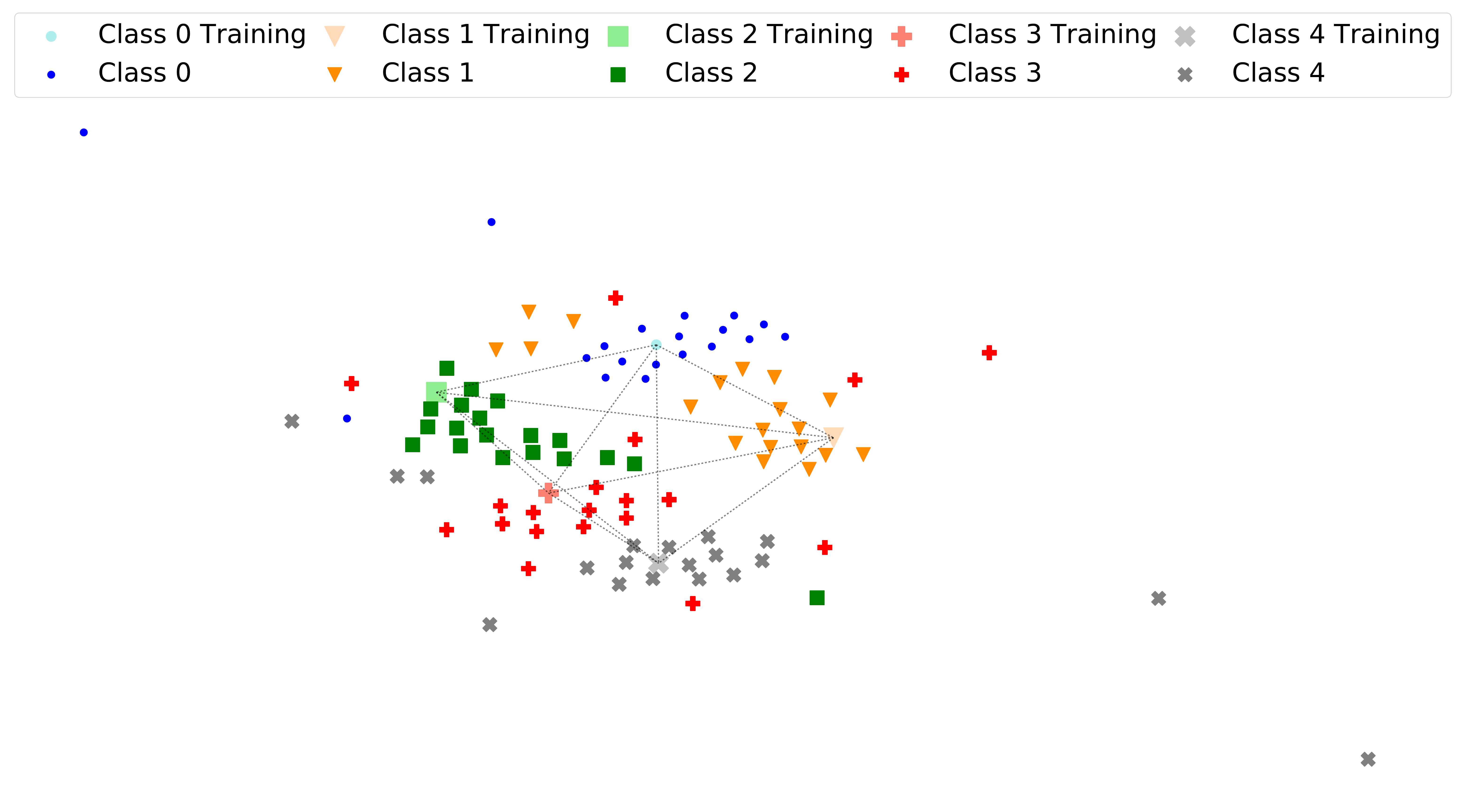}
    \caption{
         Visualization of the last hidden layer activation values by t-SNE on the Omniglot dataset before target task training (Left), and after target task training (Right). Visualized features are shown for training from scratch (Top), UMTRA (Middle), and MAML (Bottom). 
         Each class is shown by a different color and shape. From each class one instance is used for target task training. Training instances are denoted by larger and lighter symbols and are connected to each other by dotted lines
    }
    
    \label{fig:omniglot_tsne}
\end{figure}

\end{notincluded}
% \subsubsection{How much data we need for supervised MAML to span the manifold of the classes}
% We compare the accuracy of the method based on the variation in examples for supervised learning. All of our experiments are on the task of one-shot learning ($n = 5, K = 1$). Omniglot dataset has 20 instances for each class. In one set of our experiments, we perform supervised meta-learning on different number of examples each class. From 20 instances of each class, we just pick $m$ examples and we shuffled them and use them for train and validation set generation. Our experiments show that the variety of samples is an important factor in improving the accuracy of the model.

% In the other section of our experiments, we evaluate the accuracy of the model by decreasing the number of classes for training. In other words, we want to see the effect of variation between number of classes on meta-learning. We use all samples and shuffle them for each class, however, we limit the number of training set classes.

\subsection{UMTRA on the Mini-Imagenet dataset}
% (Siavash: Please read this section and fix grammatical and other errors)

The Mini-Imagenet dataset was introduced by \cite{ravi2016optimization} as a subset of the ImageNet dataset~\cite{deng2009imagenet}, suitable as a benchmark for few-shot learning algorithms. The dataset is limited to 100 classes, each with 600 images. We divide our dataset into train, validation and test subsets according to the experimental protocol proposed by~\cite{vinyals2016matching}. The classifier network is similar to the one used in~\cite{finn2017model}.

% \begin{figure}
%     \centering
%     \includegraphics[width=0.5\columnwidth]{images/augmentation/MiniImageNetAugmentationMethods.pdf}
%     \caption{Several augmentation functions used on Mini-Imagenet dataset. Auto Augment~\cite{cubuk2018autoaugment} applies augmentations from a learned policy based on combinations of translation, rotation, or shearing.}
%     \label{fig:AugmentedMiniImageNet}
% \end{figure}

Since Mini-Imagenet is a dataset with larger images and more complex classes compared to Omniglot, we need to choose augmentation functions suitable to the model. We had investigated several simple choices involving random flips, shifts, rotation, and color changes. In addition to these hand-crafted algorithms, we also investigated the learned auto-augmentation method proposed in~\cite{cubuk2018autoaugment}. Table~\ref{tab:AugmentationCompare}, right, shows the accuracy results for the tested augmentation functions. We found that auto-augmentation provided the best results, thus this approach was used in the remainder of the experiments.

The last four columns of Table~\ref{tab:ResultsOmniglot} lists the experimental results for few-shot classification learning on the Mini-Imagenet dataset. Similar to the Omniglot dataset, UMTRA performs better than learning from scratch and all the approaches that use unsupervised representation learning. It performs weaker than supervised meta-learning approaches that use labeled data. 
Compared to the various combinations involving the CACTUs unsupervised meta-learning algorithm, UMTRA performs better on 5-way one-shot classification, while it is outperformed by the CACTUs-MAML with DeepCluster combination for the 5, 20 and 50 shot classification. 

A possible question might be raised whether the improvements we see are due to the meta-learning process or due to the augmentation enriching the few shot dataset. To investigate this, we performed several experiments on Omniglot and Mini-Imagenet by training the target tasks from scratch on the augmented target dataset. For 5-way, 1-shot learning on Omniglot the accuracy was: training from scratch 52.5\%, training from scratch with augmentation 55.8\%, UMTRA 83.8\%. For MiniImagenet the numbers were: from scratch without augmentation 27.6\%, from scratch with augmentation 28.8\%, UMTRA 39.93\%. We conclude that while augmentation does provide a (minor) improvement on the target training by itself, the majority of the improvement shown by UMTRA is due to the meta-learning process.

% \noindent{\bf In particular, a simple baseline would be to perform few-shot training on models trained with the augmentation methods proposed. That seems like a more informative/reasonable baseline than training from scratch.
% }

The results on Omniglot and Mini-Imagenet allow us to draw the preliminary conclusions that unsupervised meta-learning approaches like UMTRA and CACTUs, which generate meta tasks $\mathcal{T}_i$ from the unsupervised training data tend to outperform other approaches for a given unsupervised training set $\mathcal{U}$. UMTRA and CACTUs use different, orthogonal approaches for building $\mathcal{T}$. UMTRA uses the statistical likelihood of picking different classes for the training data of $\mathcal{T}_i$ in case of $K=1$ and large number of classes, and an augmentation function $\mathcal{T}$ for the validation data. CACTUs relies on an unsupervised clustering algorithm to provide a statistical likelihood of difference and sameness in the training and validation data of $\mathcal{T}_i$. Except in the case of UMTRA with $\mathcal{A} = \mathbbm{1}$, both approaches require domain specific knowledge. The choice of the right augmentation function for UMTRA, the right clustering approach for CACTUs, and the other hyperparameters (for both approaches) have a strong impact on the performance.

% In our experiments, we found that UMTRA performs better on Omniglot compared to the best performing ACAI-CACTUs-MAML combination. On the Mini-Imagenet dataset, the best performing CACTUs combination was DeepCluster-CACTUs-MAML, compared with which UMTRA was better on the five-way one-shot learning task, while CACTUs performed better on Mini-Imagenet on (5,5), (20,1) and (20,5) tasks. 

% \input{sections/ExperimentsImages}

%\subsection{Discussion: UMTRA on few shot learning benchmarks}

%In this section, we compare UMTRA on the Omniglot and Mini-Imagenet few shot-learning benchmarks with a fixed classifier architecture. Our primary aim of comparison is with approaches which require at most an unlabeled dataset $\mathcal{U}$ for pre-training. We found that our approach outperforms, on both benchmarks and for every test combination training from scratch approaches, and every approach based on unsupervised representation learning on $\mathcal{U}$. 

%We found that our approach underperforms the control approach that used a labeled datastep for the pre-training step. 

% \input{sections/ExpUCF101}
% \input{sections/ExpCeleba}

\section{Conclusions}
\label{sec:Conclusions}

In this paper, we described the UMTRA algorithm for few-shot and one-shot learning of classifiers. UMTRA performs meta-learning on an {\bf unlabeled dataset} in an unsupervised fashion, without putting any constraint on the classifier network architecture. Experimental studies over the few-shot learning image benchmarks Omniglot and Mini-Imagenet show that UMTRA outperforms learning-from-scratch approaches and approaches based on unsupervised representation learning. 
It alternated in obtaining by best result with the recently proposed CACTUs algorithm that takes a different approach to unsupervised meta-learning by applying clustering on an unlabeled dataset. The statistical sampling and augmentation performed by UMTRA can be seen as a cheaper alternative to the dataset-wide clustering performed by CACTUs. The results also open the possibility that these approaches might be orthogonal, and in combination might yield an even better performance. For all experiments, UMTRA performed worse than the equivalent supervised meta-learning approach - but requiring 3-4 orders of magnitude less labeled data. The supplemental material shows that UMTRA is not limited to image classification but it can be applied to other tasks as well, such as video classification. 

\bigskip

\noindent{\bf Acknowledgements: } This research is based upon work supported in parts by the National Science Foundation under Grant numbers IIS-1409823 and IIS-1741431 and Office of the Director of National Intelligence (ODNI), Intelligence Advanced Research Projects Activity (IARPA), via IARPA R\&D Contract No. D17PC00345. The views, findings, opinions, and conclusions or recommendations contained herein are those of the authors and should not be interpreted as necessarily representing the official policies or endorsements, either expressed or implied, of the NSF, ODNI, IARPA, or the U.S. Government. The U.S. Government is authorized to reproduce and distribute reprints for Governmental purposes notwithstanding any copyright annotation thereon.

% This work had been supported in part by the National Science Foundation under grant numbers IIS-1409823 and
% IIS-1741431. Any opinions, findings, and conclusions or recommendations expressed in this material are those of the
% authors and do not necessarily reflect the views of the National Science Foundation.

\clearpage
\subsection*{Supplementary Material for Unsupervised Meta-Learning for Few-Shot Image Classification}
\subsubsection*{Evolution of accuracy during training}

\begin{figure}[ht]
    \centering
    \includegraphics[width=0.5\columnwidth]{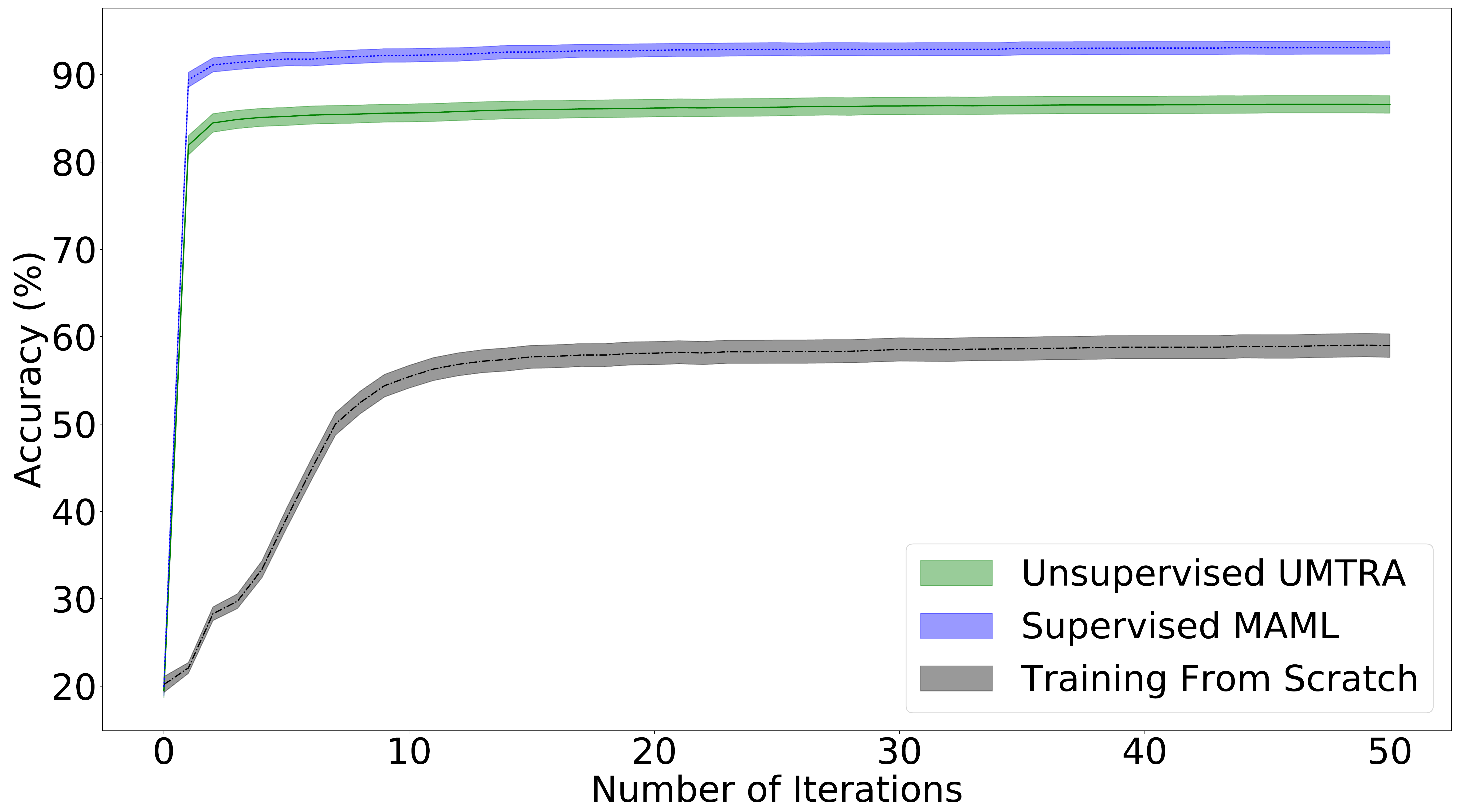}
    \caption{
        The accuracy curves during the target training task on the Omniglot dataset for $K = 1$. The band around lines denotes a $95\%$ confidence interval.
    }
    \label{fig:OmniglotAccuracyCurve}
\end{figure}

\begin{figure}
    \centering
    $K = 1$ \hspace{0.45\columnwidth} $K = 5$
    \\
    \includegraphics[width=0.49\columnwidth]{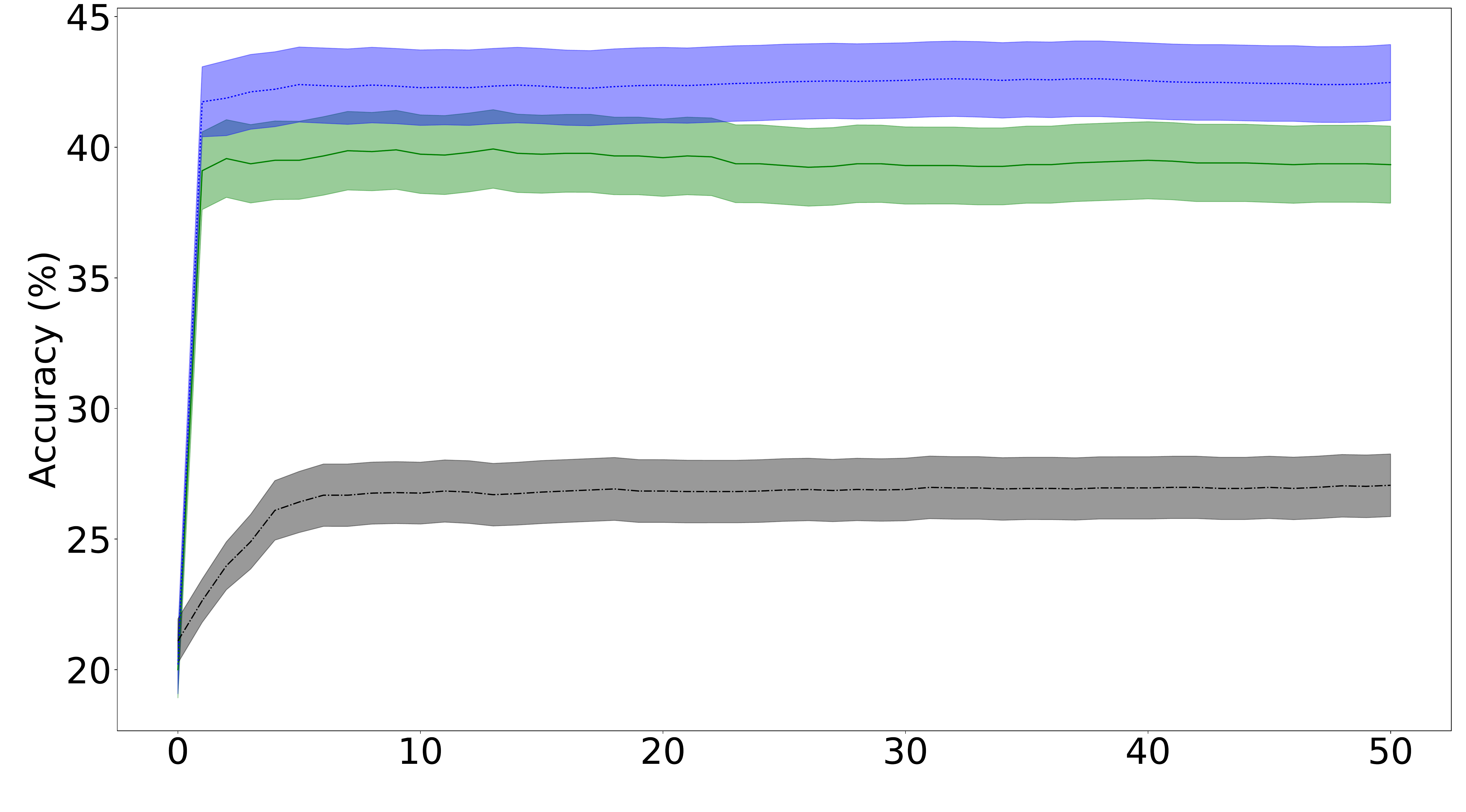}
    \includegraphics[width=0.49\columnwidth]{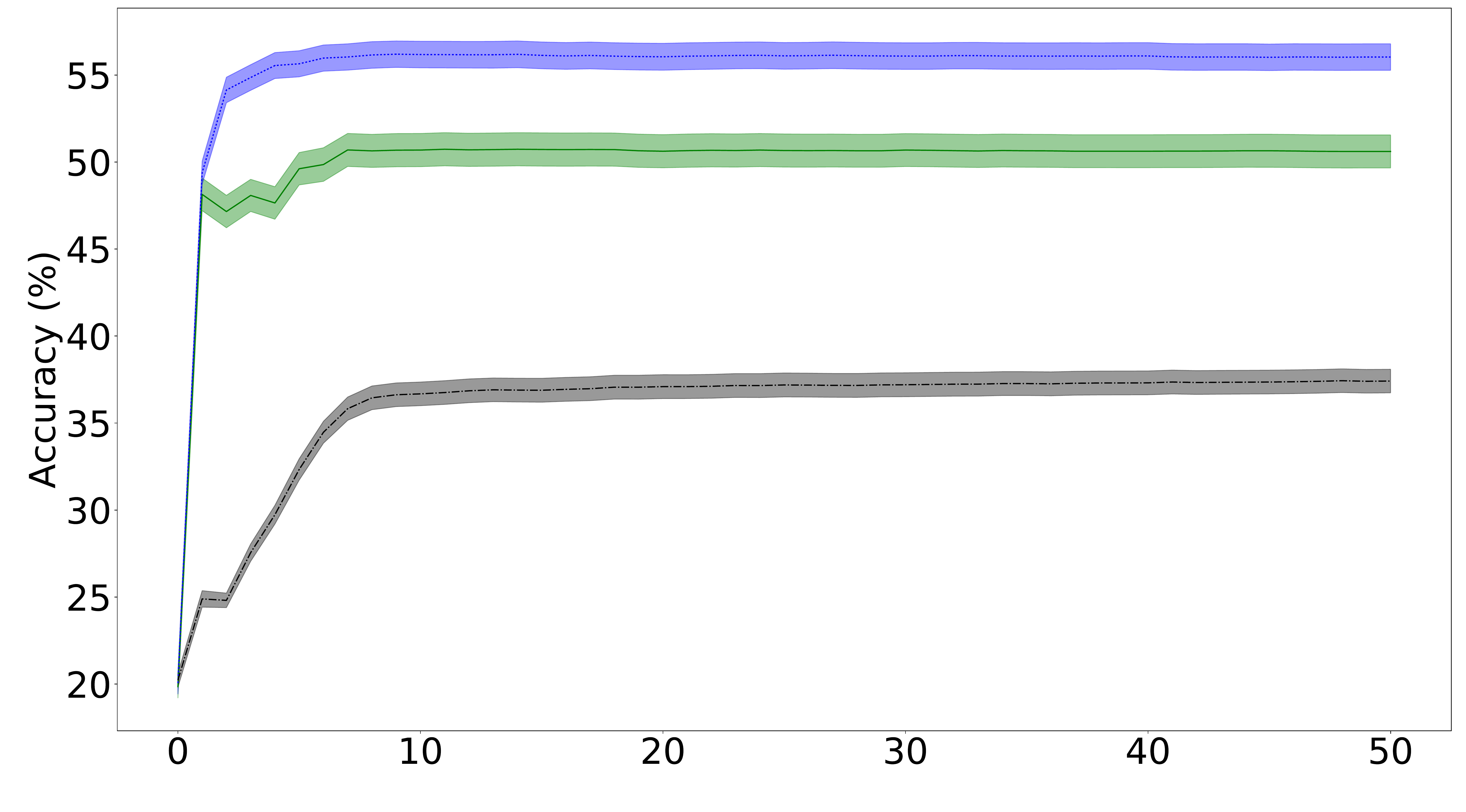}
    \\
    $K = 20$
    \\
    \includegraphics[width=0.5\columnwidth]{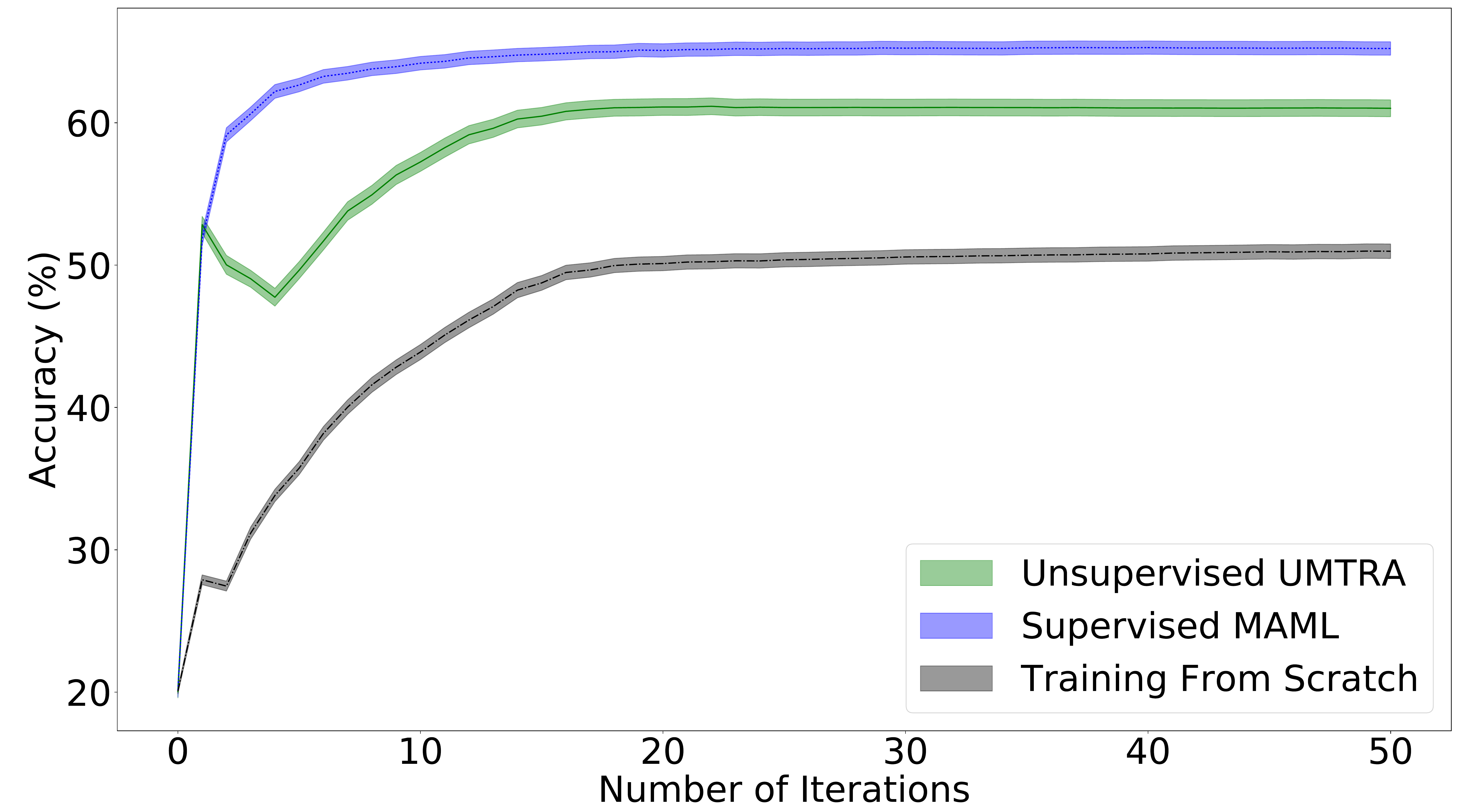}
    \caption{
        The accuracy curves during the target training task on the Mini-Imagenet dataset. Accuracy curves are shown for $K = 1$ (Top left), $K = 5$ (Top right), and $K = 20$ (Bottom).
        The band around lines denotes a $95\%$ confidence interval.
    }
    \label{fig:miniimgenetAccuracyCurve}
\end{figure}

In these series of experiments we study the evolution of the accuracy obtained after a specific number of gradient training steps during the target learning phase. The results for Omniglot are shown in Figure~\ref{fig:OmniglotAccuracyCurve} (with K=1), while those for Mini-Imagenet in Figure~\ref{fig:miniimgenetAccuracyCurve} with K values of 1, 5 and 20. For both datasets, we compare learning from scratch, UMTRA and supervised MAML. As expected, both MAML and UMTRA reach their accuracy  plateau very quickly during target training, while learning from scratch takes a larger number of training steps.
Further training does not appear to provide any benefit for either approach. The results are averaged among 1000 tasks. This demonstrates that UMTRA has the capacity to learn to adapt to novel tasks by just looking at unlabeled data and generating tasks from that dataset in an unsupervised manner.
% For neither approaches appear that further training would provide any benefit. 
An interesting phenomena happens with $K=5$ and $K=20$ values for Mini-Imagenet: the accuracy curve of UMTRA dips after the first iteration, and it takes several iterations to recover. We conjecture that this is a result of the fact that UMTRA sets $K=1$ during meta-learning, thus the resulting network is best optimized to learn from one sample per class. 

\subsubsection*{Feature Representations}
To compare generalization of training from scratch, UMTRA and supervised MAML, we visualize the activations of the last hidden layer of the network on Omniglot dataset by t-SNE. We compare all of the methods on the same target training task which is constructed by sampling five characters from test data and selecting one image from each character class randomly. Each character has 20 different instances. Figure~\ref{fig:omniglot_raw_pixel_tsne} shows the t-SNE visualization of raw pixel values of these 100 images. Instances which are sampled for the one-shot learning task are connected to each other by dotted lines. Figure~\ref{fig:omniglot_tsne} shows the visualization of the last hidden layer activations for the same task. UMTRA as well as MAML can adapt quickly to a feature space which has a better generalization than training from scratch.

\begin{figure}[ht]
    \centering
    \includegraphics[width=0.5\columnwidth]{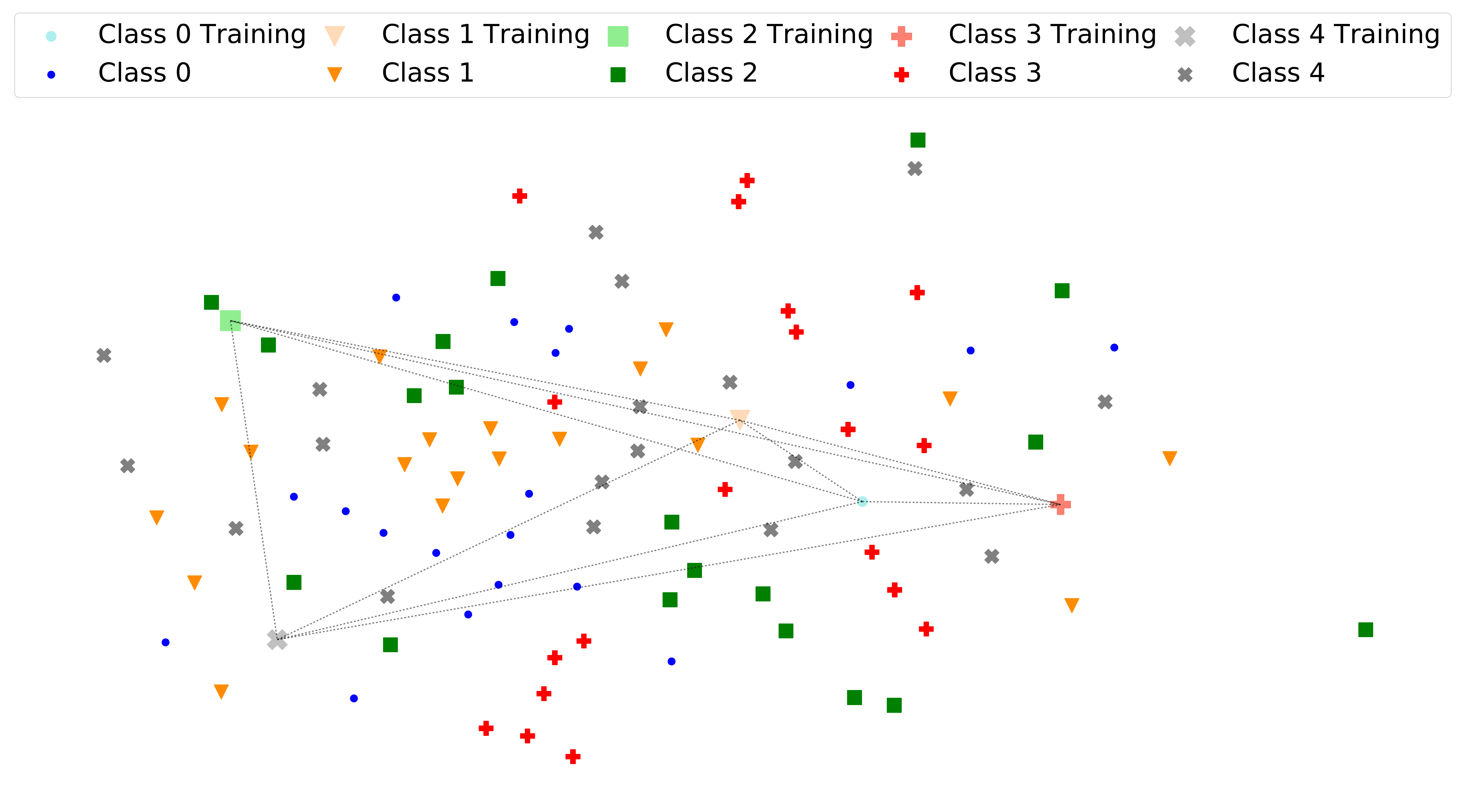}
    \caption{
        t-SNE on the Omniglot raw pixel values.
        % Training instances are denoted by larger and lighter symbol and are connected to each other by dotted lines
        % The target task contains 5 different classes. The class instances are visualized with different colors and shapes. The five instances used for target task training are connected to each other with dotted lines.
    }
    \label{fig:omniglot_raw_pixel_tsne}
\end{figure}

\begin{figure}
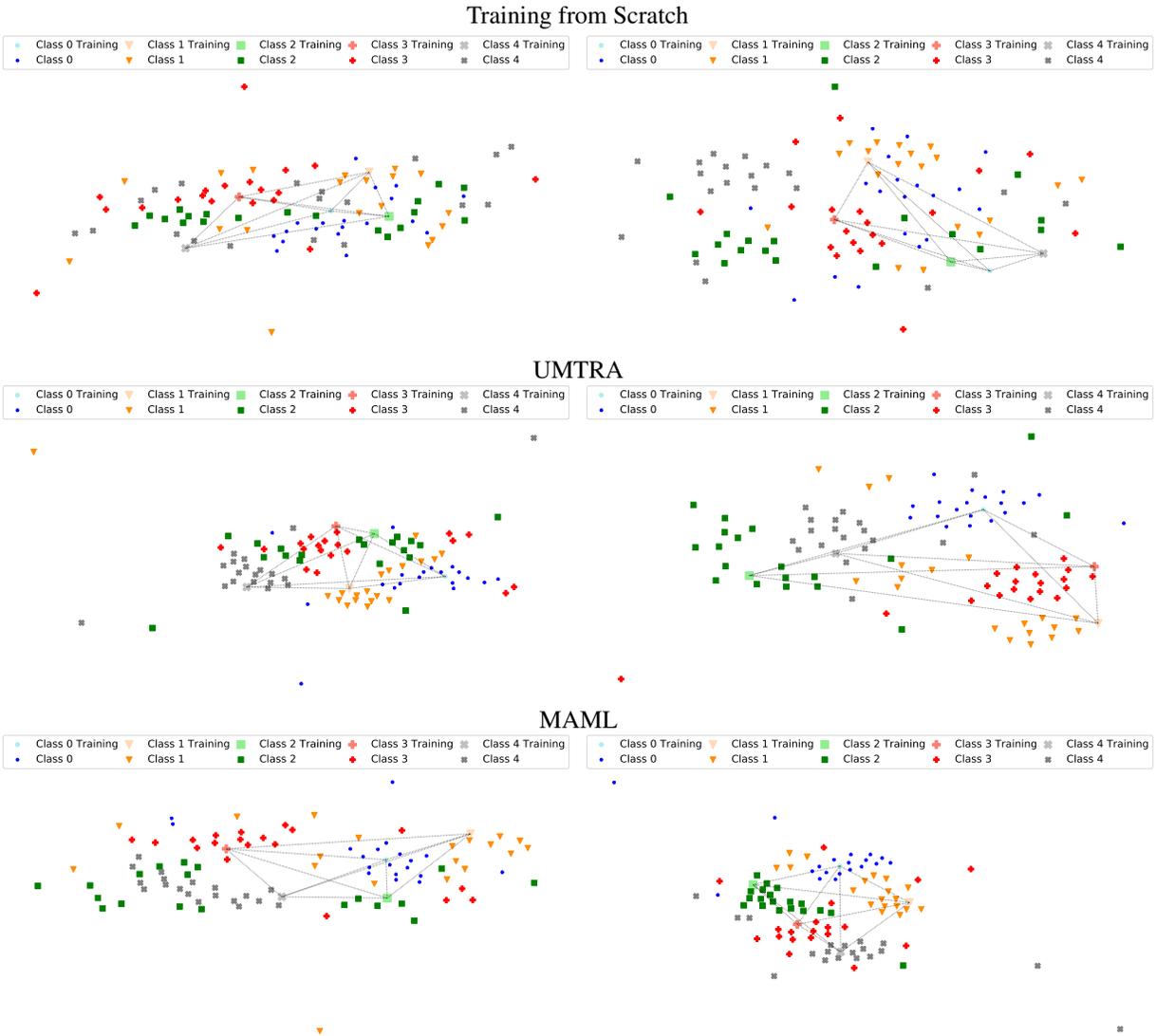

    \centering
    
    Training from Scratch
    
    \includegraphics[width=0.49\columnwidth]{images/supmat/tsne/scratch.pdf}
    \includegraphics[width=0.49\columnwidth]{images/supmat/tsne/scratchtarget.pdf}
    
    UMTRA
    
    \includegraphics[width=0.49\columnwidth]{images/supmat/tsne/unsupervised2.pdf}
    \includegraphics[width=0.49\columnwidth]{images/supmat/tsne/unsupervisedtarget2.pdf}
    
    MAML
    
    \includegraphics[width=0.49\columnwidth]{images/supmat/tsne/metalearn1.pdf}
    \includegraphics[width=0.49\columnwidth]{images/supmat/tsne/metatarget3.pdf}
    \caption{
         Visualization of the last hidden layer activation values by t-SNE on the Omniglot dataset before target task training (Left), and after target task training (Right). Visualized features are shown for training from scratch (Top), UMTRA (Middle), and MAML (Bottom). 
         Each class is shown by a different color and shape. From each class one instance is used for target task training. Training instances are denoted by larger and lighter symbols and are connected to each other by dotted lines
    }
    
    \label{fig:omniglot_tsne}
\end{figure}

\subsubsection*{Video Domain}
In this section, we show how the UMTRA can be applied to video action recognition, a domain significantly more complex and data intensive than the one used in the few-shot learning benchmarks such as Omniglot and Mini-Imagenet. To the best of our knowledge, we are the first to apply meta-learning to video action recognition. We perform our comparisons using one of the standard video action recognition datasets, UCF-101
% \cite{soomro2012ucf101}
.  UCF-101 includes 101 action classes divided into five types: Human-Object Interaction, Body-Motion Only, Human Human Interaction, Playing Musical Instruments and Sports. The dataset is composed of snippets of Youtube videos. Many videos have poor lighting, cluttered background and severe camera motion. As the classifier on which to apply the meta-learning process, we use a 3D convolution network, C3D
% ~\cite{tran2015learning}
.

Performing unsupervised meta-learning on video data, requires several adjustments to the UMTRA workflow, with regards to the initialization of the classifier, the split between meta-learning data and testing data, and the augmentation function. 

First, networks of the complexity of C3D cannot be learned from scratch using the limited amount of data available in few-shot learning. In the video action recognition research, it is common practice to start with a network that had been pre-trained on a large dataset, such as Sports-1M dataset
% ~\cite{karpathy2014large}
, an approach we also use in all our experiments. 

Second, we make the choice to use two different datasets for the meta-learning phase (Kinetics)
% ~\cite{tran2015learning, jia2014caffe, KarpathyCVPR14}
and for the few-shot learning and evaluation (UCF-101
% ~\cite{soomro2012ucf101}
). This gives us a larger dataset for training since Kinetics contains 400 actions, but it introduces an additional challenge of domain-shift: the network is pre-trained on Sports-1M, meta-trained on Kinetics and few-shot trained on UCF-101. This approach, however, closely resembles the practical setup when we need to do few-shot learning on a novel domain. When using the Kinetics dataset, we limit it to 20 instances per class.

\begin{figure}
    \centering
    \includegraphics[width=\columnwidth]{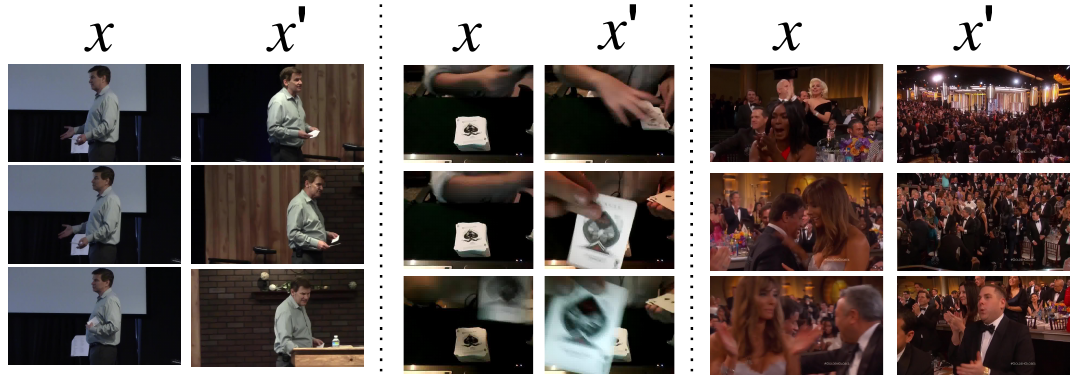}
    \caption{Example of the training data and the augmentation function $\mathcal{A}$ for video. The training data $x$ is a 16 frame segment starting from a random time in the video sample (Here we show three frames of a sample at each column). The validation data $x' = \mathcal{A}(x)$ is also a 16 frame segment, starting from a different, randomly selected time {\em from the same video sample}.}
    \label{fig:AugmentedKinetics}
\end{figure}

For the augmentation function $\mathcal{A}$, working in the video domain opens a new possibility, of creating an augmented sample by choosing a temporally shifted video fragment from the same video.
In other words, we can use self supervision in video domain: The augmentation is to sample another part of the same video clip.
Figure \ref{fig:AugmentedKinetics} shows some samples of these augmentations. In our experiments, we have experimented both with UMTRA (using a Kinetics dataset stripped from labels), and supervised meta-learning (retaining the labels on Kinetics). 
% for the choice of the validation, but following the rest of the experimental protocol). 
This supervised meta-learning experiment is also significant because, to the best of our knowledge, meta-learning has never been applied to human action recognition from videos. 

In our evaluation, we perform 30 different experiments. At each experiment we sample 5 classes from UCF-101, perform the one-shot learning, and evaluate the classifier on all the examples for the 5 classes from UCF-101. As the number of samples per class are not the same for all classes, in  Table~\ref{tab:UCF101-Results} we report both the accuracy and F1-score.

The results allow us to draw several conclusions. The relative accuracy ranking between training from scratch, pre-training and unsupervised meta-learning and supervised meta-learning remained unchanged. Supervised meta-learning had proven feasible for one-shot classifier training for video action recognition. UMTRA performs better than other approaches that use unsupervised data. Finally, we found that the domain shift from Kinetics to UCF-101 was successful. 

\begin{table}
\caption{Accuracy and F1-Score for a 5-way, one-shot classifier trained and evaluated on classes sampled from UCF-101. All training (even for ``training from scratch''), employ a C3D network pre-trained on Sports-1M. For all approaches, none of the UCF-101 classes was seen during pre- or meta-learning.}
    \label{tab:UCF101-Results}
    \centering
    {\footnotesize
    \begin{tabular}{p{5cm}p{3.5cm}}
    \hline
    {\bf Algorithm} & {\bf Test Accuracy / F1-Score} \\
        \hline 
        {\em Training from scratch} & 29.30 / 20.48\\
        \hline
        Pre-trained on Kinetics & 45.51 / 42.49\\ 
        \hline
        UMTRA on unlabeled Kinetics (ours) & \textbf{60.33 / 58.47}\\
        \hline
        Supervised MAML on Kinetics & 71.08 / 69.44\\
        \hline
        \end{tabular}
    }
    \vspace{1mm}
\end{table}
% \twocolumn
% \input{Supplementary.tex}

\end{document}